\PassOptionsToPackage{hyphens}{url}
\documentclass[a4paper, 12pt, nonatbib]{elsarticle}
\makeatletter
\def\ps@pprintTitle{%
	\let\@oddhead\@empty
	\let\@evenhead\@empty
	\def\@oddfoot{\centerline{\thepage}}%
	\let\@evenfoot\@oddfoot}
\makeatother

\usepackage{latexsym, amssymb, amsmath, amsfonts, amsthm, bm, bbm, nicefrac, stmaryrd, thmtools, enumitem, centernot}
\usepackage{url, graphicx, xcolor, rotating}
\definecolor{svlinks}{rgb}{.0,0.3,0.6}
\usepackage[hang]{footmisc}
\usepackage[bookmarks,colorlinks=true,pdfhighlight=/O,linkcolor=svlinks,urlcolor=svlinks,citecolor=svlinks,
           pdftitle={k-Anonymity in Practice: How Generalisation and Suppression Affect Machine Learning Classifiers},
           pdfauthor={Djordje Slijepčević, Maximilian Henzl, Lukas Daniel Klausner, Tobias Dam, Peter Kieseberg, Matthias Zeppelzauer},
           pdfsubject={},
           pdfkeywords={}]{hyperref}
\usepackage[T2A,T1]{fontenc}
\usepackage[utf8]{inputenc}
\usepackage[russian,greek,UKenglish]{babel}
\usepackage[normalem]{ulem}

\usepackage{epsfig, listings, appendix}
\usepackage{multirow}
\usepackage{float}
\usepackage{placeins}
\usepackage[all]{nowidow}
\usepackage{fnpct}

\usepackage{tikz, wrapfig, caption}
\usetikzlibrary{matrix}
\usepackage[normalem]{ulem}
\usepackage{booktabs}
\usepackage{array}
\usepackage[subfolder]{gnuplottex}
\usepackage{svg}

\definecolor{clr}{RGB}{30, 200, 30}

\setlist{nosep}

\usepackage{etoolbox}
\pretocmd{\eqref}{Eq.~}{}{}

\usepackage{todonotes}

\def\eg{e.\,g.}
\def\ie{i.\,e.}

\makeatletter
\let\c@author\relax
\makeatother

\usepackage[style=authoryear,sorting=nyt,sortcites=true,autopunct=true,babel=hyphen,hyperref=true,abbreviate=false,backref=false,backend=biber]{biblatex}
\bibliography{Bibliography}
\begin{document}

\begin{frontmatter}

\title{$k$-Anonymity in Practice: How Generalisation and Suppression Affect Machine Learning Classifiers}

\author[ICMT]{Djordje Slijep\v{c}evi\'c\textsuperscript{*}\textsuperscript{\textsection}}
\ead{djordje.slijepcevic@fhstp.ac.at}

\author[ISR]{Maximilian Henzl\textsuperscript{\textsection}}
\ead{is201849@fhstp.ac.at}

\author[ISR]{Lukas Daniel Klausner}
\ead{mail@l17r.eu}

\author[ISR]{Tobias Dam}
\ead{tobias.dam@fhstp.ac.at}

\author[JRC]{Peter Kieseberg}
\ead{peter.kieseberg@fhstp.ac.at}

\author[ICMT]{Matthias Zeppelzauer}
\ead{matthias.zeppelzauer@fhstp.ac.at}

\address[ICMT]{
Institute of Creative\textbackslash Media/Technologies, St.\ P\"olten University of Applied Sciences, 
Austria
}

\address[ISR]{
Institute of IT Security Research, 
St.\ P\"olten University of Applied Sciences,
Austria
}

\address[JRC]{
Josef Ressel Center for Blockchain Technologies and Security Management,
St.\ P\"olten University of Applied Sciences, Austria
}

\begin{abstract}

    The protection of private information is a crucial issue in data-driven research and business contexts. Typically, techniques like anonymisation or (selective) deletion are introduced in order to allow data sharing, \eg\ in the case of collaborative research endeavours. For use with anonymisation techniques, the $k$-anonymity criterion is one of the most popular, with numerous scientific publications on different algorithms and metrics. Anonymisation techniques often require changing the data and thus necessarily affect the results of machine learning models trained on the underlying data. In this work, we conduct a systematic comparison and detailed investigation into the effects of different $k$-anonymisation algorithms on the results of machine learning models. We investigate a set of popular $k$-anonymisation algorithms with different classifiers and evaluate them on different real-world datasets. Our systematic evaluation shows that with an increasingly strong $k$-anonymity constraint, the classification performance generally degrades, but to varying degrees and strongly depending on the dataset and anonymisation method. Furthermore, Mondrian can be considered as the method with the most appealing properties for subsequent classification.
\end{abstract}


\end{frontmatter}

\begingroup\renewcommand\thefootnote{\textsection}
\footnotetext{These authors contributed equally to the article.}
\endgroup

\section{Introduction}

The amount of human-generated data that is being stored, processed and analysed is growing exponentially. A significant part of these data is by nature personal and sensitive information, \eg\ names, location data, ethnicity, health condition, political opinions, gender identity or sexual orientation. Such information is collected in different contexts and is used by companies as well as research institutions, governmental and non-governmental organisations. Data-driven research and products incorporating machine learning (ML) methods rely on the (automated) analysis of personal information to generate knowledge or provide (personalised) services. Consequently, the use, distribution and publication of such data poses many challenges. The main challenge is to ensure the protection of privacy of individuals whose data are being stored and processed while maintaining the usability of the data.

Privacy is considered a fundamental human right and is therefore protected to varying degrees by a multitude of different national and regional laws. In the European Union, the General Data Protection Regulation (GDPR, EU 2016/679)~\parencite{regulation2016regulation} provides a very strict and compulsory framework for the protection of personal and sensitive information. While the ``privacy by design'' paradigm~\parencite{langheinrich2001privacy} stresses the precept of using as little sensitive information as possible, many applications directly rely on the collection and processing of personal information. Thus, in order to comply with privacy protection legislation (including the GDPR) the anonymisation of sensitive data is absolutely essential.

Anonymisation aims at ensuring that data records of a person can no longer be unambiguously traced back to this specific person. For this purpose, a variety of competing and complementary privacy paradigms have been defined, \eg\ $k$-anonymity~\parencite{samarati2001protecting}, $\ell$-diversity~\parencite{machanavajjhala2007l}, $t$-closeness~\parencite{li2007t}, $\delta$-pres\-ence~\parencite{nergiz2007hiding}, and $\varepsilon$-differential privacy~\parencite{dwork2006calibrating}. The most popular privacy model for protecting privacy in data is $k$-anonymity~\parencite{gkoulalas2014publishing}. In order to satisfy $k$-anonymity, it is sufficient to transform the \textit{quasi-identifiers} (QIDs), \ie\ attributes such as location information, age, ethnicity or gender that could be used (in combination with external information) to re-identify individuals. This transformation must ensure that each record shares the same values with at least $k-1$ other records in the dataset. However, even the application of this rather simple anonymity paradigm distorts the data. This can result in information loss and thus may introduce a bias for ML models. The resulting changes to the ML results are difficult to estimate in advance. The magnitude of the data distortion and the resulting information loss are decisive for the usability of the underlying data. Keeping the information loss as small as possible is crucial, especially for automated analysis by means of ML methods, which aims to derive meaningful patterns from the underlying data. The reduced usability is also a reason why we focus primarily on $k$-anonymity and do not include other privacy paradigms mentioned above such as \eg~$\varepsilon$-differential privacy, which is a competing anonymisation paradigm that is often used in practice but seems unable to provide acceptable ML utility for values of $\varepsilon$ that provide practically relevant protection~\parencite{domingo2021limits}.

$k$-anonymity implies a manipulation of the values of QIDs for which several strategies exist~\parencite{gkoulalas2014publishing}: \textit{generalisation}~\parencite{samarati2001protecting}, \textit{suppression}~\parencite{samarati2001protecting}, \textit{microaggregation}~\parencite{domingo2002practical} and \textit{bucketisation}~\parencite{xiao2006anatomy}. A variety of algorithms aiming to achieve $k$-anonymity have been proposed~\parencite{ciriani2008k}. However, the selection of the most suitable anonymisation algorithm is challenging in itself and even more so when the aim is to apply ML on the anonymised data. Previous work introducing novel anonymisation algorithms often neglected the relations between anonymisation and ML, which may partly be due to the fact that these two methodologies originate from different domains. Thus, the literature is sparse on comparison of general-purpose anonymisation approaches investigating their advantages in the light of ML. One valuable contribution is the systematic comparison carried out by \textcite{AyalaRivera2014}. The authors addressed the issue of information loss for generalisation and suppression algorithms (although from an information-theoretic perspective and not from the perspective of applied ML). To this end, the authors employed Generalised Information Loss as the metric for comparison. \textcite{rodriguez2018does} conducted a systematic comparison of anonymisation algorithms which employ microaggregation with respect to their influence on ML. For this purpose, they used two real-world and two artificially generated datasets and investigated exclusively binary classification tasks. Their results showed that microaggregation methods have no significant influence on classification performance. However, since no such systematic comparison has been conducted for anonymisation methods employing generalisation and suppression, the question arises whether a similar or completely contrary result may be observed for these methods. 

The primary aim of this article is to investigate the influence of general-purpose $k$-anonymisation algorithms employing generalisation and suppression on ML results and to thereby fill a gap in the current literature. Although \textcite{Malle2017} already presented a preliminary analysis in which they examined the influence of the greedy clustering algorithm SaNGreeA~\parencite{campan2008data} on the classification of the \textsc{adult} dataset (see \autoref{sub:datasets} below for information on datasets), this problem requires a more comprehensive analysis. For this purpose, we conduct a systematic comparison that extends the scope of evaluations performed until today. We examine a comprehensive set of parameters and settings: four general-purpose $k$-anonymisation algorithms (\eg\ Mondrian~\parencite{LeFevre2006} and Optimal Lattice Anonymization~\parencite{ElEmam2009}), different privacy requirements (\ie\ different values for $k$ and different suppression values) as well as four real-life datasets. Through experimental evaluation of common evaluation metrics (\ie\ classification accuracy, precision, recall and F\textsubscript{1}~score), we demonstrate how strong the effects of anonymisation are on the final ML results and which anonymisation algorithms are most suitable for the subsequent ML task. 

The main results from our systematic comparison show that:
\begin{itemize}
    \item With an increasingly strong $k$-anonymity constraint, classification performance generally degrades (with the degree of degradation strongly dependent on dataset and anonymisation method).
    \item For some datasets, even for very large $k$ of up to 100 (which is far higher than the values used in practice nowadays), the performance loss remains within acceptable limits.
    \item Mondrian provides a better basis for subsequent classification than other anonymisation algorithms investigated, \ie~Optimal Lattice Anonymization, Top-Down Greedy Anonymisation and $k$-NN Clustering-Based Anonymisation.
\end{itemize}

The article is organised as follows: ``Related Work'' reviews similar previous work in the literature, ``Methodology'' describes the anonymisation and ML techniques investigated in our study, and ``Study Design'' presents the experimental setup and the datasets used. In ``Experimental Results'', we present and discuss the results obtained, before concluding the article with our main findings and presenting several possibilities for expanding our work in the conclusion.

\section{Related Work}

The theoretical study of the effects of $k$-anonymisation and related privacy models on data quality has mostly been focused on the analysis of \textit{information metrics} of the data itself. \textcite{Fung2010} give a comprehensive overview of privacy models, anonymisation algorithms and information metrics; despite its publication date (2010), this article continues to be one of the most extensive surveys on the topic.

Relatively few studies comparable to our methodology (\ie\ focusing on how anonymisation affects ML performance) have been published so far. Existing studies have mostly focused on the development of novel and more efficient anonymisation algorithms (in contrast to our open-ended approach). Three prominent examples are the following:
\begin{itemize}
    \item \textcite{Fung2007} proposed a novel anonymisation algorithm called \textit{Top-Down Refinement}, which is based on information metrics (an improvement over their prior algorithm Top-Down Specialisation~\parencite{Fung2005}). The proposed algorithm is examined in terms of classification error obtained on three datasets. In addition to the \textsc{adult} dataset, which is also used in our experiments, the authors consider two smaller datasets (with 653 and 1,000 entries, respectively). The algorithm is compared to only one competing anonymisation method.
    \item \textcite{Li2011} based their algorithm \textit{Information-Based Anonymisation for Classification Given $k$} (IACk) on normalised mutual information as a metric. IACk outperforms InfoGain Mondrian~\parencite{lefevre2006workload} (another utility-aware anonymisation algorithm) in terms of classification performance using several different classifiers on the \textsc{adult} dataset.
    \item \textcite{Last2014} proposed an algorithm called \textit{Non-Homogeneous Generalisation with Sensitive Value Distribution} (NSVDist), which is based on an information loss metric. The authors perform a relatively broad comparison on eight datasets and four different classifiers and compare their proposed NSVDist method with three other anonymisation methods: Mondrian, Privacy-Aware Information Sharing (PAIS)~\parencite{mohammed2009anonymizing} and Sequential Anonymization (SeqA)~\parencite{goldberger2009efficient}. In contrast to their approach, we perform a comparative analysis of several well-established anonymisation methods and conduct a more in-depth investigation into the effects of generalisation and suppression on ML performance.
\end{itemize}

Further examples drawing a limited comparison include \textcite{Han2017} (who also base their analysis -- in which they compare their algorithm to IACk -- on the \textsc{adult} dataset), and \textcite{Silva2017} (who use their own dataset of public transport data from Curitiba, Brazil, to perform a vertical analysis of performance across different levels of anonymisation using the ARX Data Anonymization Tool\footnote{\url{https://arx.deidentifier.org/}}).

\textcite{Inan2009} follow a different approach than the previously mentioned work. Instead of anonymising existing datasets and fine-tuning either the anonymisation algorithms or the ML models to improve the results on the anonymised data, they devise a method of performing calculations (for $k$-NN and SVM classifiers) on (already) anonymised data by considering them as ``uncertain'' data and employing stochastic arguments (\ie\ working with expected values) for downstream tasks such as classification.

Even for the methodology of comparing anonymisation algorithms using information metrics, the existing literature contains only small-scale studies comparing relatively few algorithms (to name a few, \textcite{AyalaRivera2014}, \textcite{Ghinita2007}, \textcite{LeFevre2006}, \textcite{Xu2006}); and even with that restriction, comparing results between these studies turns out to be quite difficult, due to varying methodologies and lack of publicly available implementations and annotated datasets (\eg\ including definitions of QIDs and generalisation hierarchies). In this article, we provide a more comprehensive study on different anonymisation and ML approaches as well as different datasets. For reproducibility we make our implementations, datasets, annotations, QIDs, and generalisation hierarchies publicly available\footnote{\url{https://github.com/fhstp/k-AnonML}} to stimulate further research on this topic.

\section{Methodology}

In the following sections we describe the approaches investigated and compared in our study. The approaches include the anonymisation and ML techniques described below.

\subsection{Anonymisation Algorithms and Information Metrics}

\begin{figure}[h]
    \centering
    \includegraphics[width=.7\linewidth]{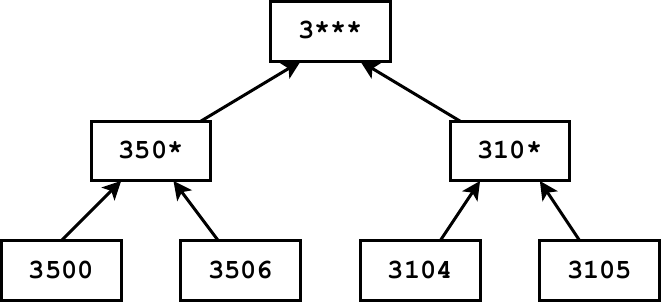}
    \caption{Value generalisation hierarchy for postal codes.} 
    \label{fig:vgh}
\end{figure}

In order to ensure a reasonable amount of comparability, all algorithms included in this study use the common principle of \emph{generalisation}. This method consists of replacing values of an attribute with a more generalised value; this is usually achieved by utilising either so-called \emph{domain generalisation hierarchies} (DGH) or \emph{value generalisation hierarchies} (VGH)~\parencite{samarati2001protecting}. A DGH describes the relationship between domains and consists of an attribute domain, all possible values of an attribute, as well as the related, more generalised domains; for instance, the domain of a postal code attribute might be \texttt{\{3500, 3506, 3104, 3105\}}, which can be generalised to \texttt{\{350*, 310*\}}. A VGH contains additional information about generalisation steps for individual attribute values; VGH are often visualised as tree structures to illustrate the relationship between values in a specific domain and according values in more generalised domains. An examplary VGH for the aforementioned postal code values is shown in \autoref{fig:vgh}.

Generalisation algorithms can be divided into two different types~\parencite{AyalaRivera2014}: global (also called \emph{full-domain generalisation}) and local algorithms. \emph{Global generalisation} applies the same generalisation step to each attribute with the according value. In our example, the postal code attribute in every data record containing \texttt{3500} is generalised to \texttt{350*}. In contrast to this approach, \textit{local generalisation} allows the generalisation of only some attributes with the same value, while other attributes with this value remain unchanged, which enables a more fine-grained processing of the data to achieve $k$-anonymity with less data distortion.

Global generalisation algorithms can be further distinguished into \emph{single-dimensional} and \emph{multi-dimensional} algorithms. The former applies the generalisation to each attribute independently, whereas the latter considers a group of attributes jointly to find a suitable generalisation of the data.

The concept of generalisation can be supplemented by additionally applying \emph{suppression}~\parencite{sweeney2002achieving}, which adds a new maximal element to the VDH (though this is sometimes omitted if there already is a unique maximal element corresponding to ``all information removed''). This element is typically represented by replacing values entirely by asterisks (\eg\ $\texttt{3500} \rightarrow \texttt{****}$), which means that all information is withheld. Suppression can be applied either to the entire data record (\ie\ \emph{record suppression}) or only to specific attributes of a record (also known as \emph{cell suppression}).

In our investigation, we focus on anonymisation algorithms employing either just generalisation or both generalisation and suppression, cf.\ \textcite[section~3, especially subsection~3.1]{Fung2010}. We base our analysis on a selection of anonymisation algorithms chosen to represent a diverse set of methodical characteristics. In particular, our choice of algorithms contains both optimal and heuristic methods as well as representatives of several common generalisation strategies (full-domain generalisation, cell generalisation and multidimensional generalisation). Additionally, we compare generaralisation and suppression with microaggregation to provide a more comprehensive picture.

\subsubsection{Optimal Lattice Anonymization}
\label{subsub:ola}
Optimal Lattice Anonymization~(OLA)~\parencite{ElEmam2009} is an optimal $k$-anonymity algorithm with record suppression which works by searching for an optimal node inside a lattice of possible generalisation steps. The lattice is a combination of the different VGHs of the underlying data and is organised into levels based on the combined generalisation level of the distinct attributes. Paths in the lattice (from the bottom to the top) correspond to generalisation strategies. The first step of the algorithm lies in performing a binary search for each such generalisation strategy in order to find all $k$-anonymous nodes (for some fixed $k$) utilising predictive tagging. Predictive tagging is used to reduce how often the algorithm has to check whether a node satisfies $k$-anonymity; this makes use of the facts that nodes above a $k$-anonymous node are $k$-anonymous and nodes below a non-$k$-anonymous node are not $k$-anonymous. The binary search algorithm is performed by iterating over nodes starting in the median height of the lattice and checking whether the node is $k$-anonymous as well as which extent of suppression is required. Depending on the result, the nodes above or below are accordingly tagged as fulfilling or violating $k$-anonymity, as well. The untagged half is then separated into a sublattice and the steps are repeated. This search process reveals the $k$-anonymity of a significant portion of nodes without requiring explicit calculations for all of them. The second step is to remove elements from the set of $k$-anonymous nodes such that only level-minimal (also called $k$-minimal) nodes within a generalisation strategy remain. Finally, all $k$-minimal nodes are compared based on the discernibility metric (DM) (or other suitable metrics) and the node with the smallest information loss is chosen as the optimal solution. OLA uses a modified version of the original DM~\parencite{BayardoAgrawal2005} which assigns a penalty to each tuple based on either how many indistinguishable tuples are contained in the anonymised table or whether the tuple is suppressed.

\subsubsection{Mondrian}
Mondrian~\parencite{LeFevre2006} is a greedy approximation algorithm for achieving $k$-anonymity by partitioning the domain space into multidimensional regions. For this study, we extended an open-source implementation of the algorithm\footnote{\url{https://github.com/qiyuangong/Basic_Mondrian}} to allow for its joint application with ML methods. Our extension includes the option to leave non-QID attributes and the target variable non-anonymised as well as the ability to handle float numbers in datasets. In addition, we ported the code to Python 3 and removed unused functions and files.
Since Mondrian is designed to work top-down, it uses the highest generalisation of the QIDs as a starting point and recursively specialises into partitions by applying multidimensional cuts until no further cuts are available. Each iteration of the algorithm needs to choose a dimension (attribute) on which to perform the cut. The general-purpose approach is to use the dimension with the widest range of values. Afterwards, the split value is determined by using median partitioning, and the cut is performed according to the split value. Mondrian can be used for strict partitioning (utilising global generalisation) as well as for relaxed partitioning (applying local generalisation). While the original Mondrian algorithm is already able to handle categorical attributes by assuming a total order of values (as for numerical attributes), LeFevre et al.\ also defined an extended version which utilises value generalisation hierarchies instead~\parencite{lefevre2006workload}.

\subsubsection{Top-Down Greedy Anonymisation}
\label{subsub:tdg}
\textcite{Xu2006} proposed a simple heuristic local generalisation method based on a top-down greedy~(TDG) approach for anonymisation. For this study, we extended an open-source implementation of this algorithm\footnote{\url{https://github.com/qiyuangong/Top_Down_Greedy_Anonymization}}. We performed the same implementation enhancements as we did for Mondrian. The algorithm takes a table containing the data as input and recursively partitions it into equivalence classes which are more and more local. For this purpose, binary partitioning in combination with a heuristic is used to bisect the data in each iteration. The normalised certainty penalty (NCP) metric plays a central role; NCP incorporates both the information loss caused by anonymisation as well as the importance of the attributes. Furthermore, NCP measures the uncertainty of the attribute values of the generalised record, comparing them with the original ones' and weighting them accordingly. The data records causing the highest NCP when put into the same equivalence class represent the starting point for the two equivalence classes: The initial tuples for each equivalence class are found by randomly picking a tuple $u$ and calculating the NCP with every other tuple $v$; the tuple $v^*$ with the highest NCP is then used as the starting point for another iteration, in which the NCP with all other tuples is calculated again. This process is repeated until the resulting NCP does not change substantially anymore, thus fixing two tuples as the basis for the bisection. The other data records of the table are then assigned to one of the two equivalence classes by minimising the NCP. After the partitioning is complete, all equivalence classes containing less than $k$ elements are postprocessed to achieve $k$-anonymity. For each such equivalence class $G$, the following two steps are applied: The first step lies in searching, within all equivalence classes of size at least $2k -|G|$,\footnote{This condition is necessary to ensure that after removing the subset $G_{s}$ from such an equivalence class $H$ of size at least $2k -|G|$, the resulting smaller equivalence class $H \smallsetminus G_{s}$ is still of size at least $k$.} for the subset $G_{s}$ of tuples of size $k - |G|$ with the lowest $NCP(G \cup G_{s})$. The increase of the penalty as a result of merging $G$ with the according nearest neighbour equivalence class is calculated, as well, and compared to the former penalty measure. The solution resulting in a lower overall penalty is applied; the whole process is then repeated until all equivalence classes have at least size $k$.

\subsubsection{$k$-NN Clustering-Based Anonymisation}
\label{subsub:cb}
Another approach for grouping data records such that the anonymised records satisfy $k$-anonymity is \emph{clustering}, which aims at partitioning records into equivalence classes of similar records~\parencite{LinWei2008}. For this study, we extended an open-source implementation of a $k$-nearest neighbour clustering-based~(CB) algorithm\footnote{\url{https://github.com/qiyuangong/Clustering_based_K_Anon}}. We performed the same implementation enhancements as we did for Mondrian. The iterative clustering procedure works similar to other $k$-NN clustering-based anonymisation approaches~\parencite{aggarwal2004condensation}. During each iteration, a record is randomly picked out of the dataset and the closest $k-1$ other records are determined via a distance function. Those records (as well as the chosen record) are assigned to one equivalence class and removed from the original dataset. This process is repeated until either all records have been processed or less than $k$ records remain; these remaining records are assigned to the nearest equivalence class. The algorithm we use employs VGHs for generalisation and the NCP (as described in \autoref{subsub:tdg}) as the distance function. The distance is computed using the generalisation of the corresponding records or clusters.

\subsection{Machine Learning Algorithms}
For the present study, we investigated popular supervised ML methods, including Support Vector Machines~(SVM), $k$-Nearest Neighbour~($k$-NN), Random Forests~(RF) and Extreme Gradient Boosting~(XGBoost). Since our motivation is to investigate the interplay of ML and anonymisation algorithms, we have chosen to eschew deep learning models and concentrate on traditional ML approaches which provide at least a modicum of transparency.

\textbf{SVMs} are popular supervised ML methods used for classification and regression. SVMs are effective and robust for high-dimensional input data (in many cases even if the number of features is greater than the number of samples). They are also versatile and flexible, as many kernel functions (\eg\ linear, polynomial, radial basis function) can be specified as decision functions, thus allowing high adaptability to the input data. In contrast to more complex kernels, linear kernels are characterised by significantly shorter runtimes and little overfitting while still yielding comparatively good results. SVMs are sensitive to hyperparameters, \eg\ the cost parameter $C$ in the case of the linear kernel.

\textbf{$k$-NN} is a simple and intuitive instance-based algorithm that does not require any actual model training process. To determine the class for a given test sample, a majority decision is made based on the class membership of a given number of nearest neighbours from the training data. To determine which neighbours are closest, a similarity metric (\eg\ Euclidean distance) is used. For a sufficiently large dataset, great results can be obtained, but with unbalanced data the algorithm encounters difficulties. The method is very sensitive to the number of neighbours used to determine the class of a tested instance.

\textbf{RFs}~\parencite{breiman2001random} are robust supervised ML methods based on an ensemble of simple decision trees. Individual decision trees are relatively inflexible and not robust because even small changes in the data can cause the generated decision trees to look very different. Building RFs involves first generating simple decision trees from different subsamples of the data and then combining the results (\eg\ by averaging their probabilistic predictions) into a relatively robust joint model. RFs are sensitive to the number of decision trees employed, \ie\ a sufficient number of decision trees is necessary to obtain robust predictions.

\textbf{XGBoost}~\parencite{chen2016xgboost} is a highly optimised and efficient variant of Gradient Boosting~\parencite{friedman2001greedy}. Similar to RFs, Gradient Boosting combines a set of decision trees to provide robust predictions. The difference lies in the generation of decision trees: RFs generate independent decision trees on random subsets of data and then combine their results, whereas Gradient Boosting learns the trees iteratively and also learns from existing trees.

\section{Study Design}

We apply the following approach for our comparison: First, we implement the general-purpose anonymisation algorithms defined in the literature and described above or use existing implementations thereof. Next, we apply them to a selection of datasets and finally, we evaluate the influence of these algorithms on the performance of the classification methods described above. We use all combinations of anonymisation algorithms, ML methods and datasets to obtain the most complete picture possible of the dependencies between these components and to provide the basis for an objective comparison.

\subsection{Experimental Setup}

\paragraph{Pre-processing and baselines} The first step includes the pre-processing of the data, \ie\ defining the generalisation hierarchies and performing one-hot encoding of categorical QIDs. In our experiments, we use only QIDs and the target variable while removing all other attributes to ensure that additional attributes that may be highly correlated with the target variable do not bias the results. For numerical QIDs we opted to use the mean value of the generalised interval to maintain $k$-anonymity. The data is randomly split into training~(70\%) and test~(30\%) sets. To establish the baselines, each of the investigated classifiers is initialised and then trained on the training set. The trained model is used to predict the target variable on the test set. Thereby, a non-anonymised baseline is calculated for the classification model. Additionally, we determined the zero-rule baseline~(ZRB) as the lower baseline in our study by evaluating a naive classifier which always predicts the most frequent class in the test set. We fixed the hyperparameters for each classifier across all experiments: For the linear SVM $C=1$, for $k$-NN the number of neighbours $k$ is set to 10, and for RF and XGBoost the number of individual decision trees is set to 300 and 100, respectively. 

\paragraph{Anonymisation} After establishing the baselines, the respective datasets are anonymised according to the chosen parameters (\ie\ the used algorithm and the value of $k$). The anonymised data are then split (with the same training/test split used in the non-anonymised setting) and used to train the model and predict the target variable as before. This is repeated for different anonymisation algorithms and parameters (multiple values of $k$ and degrees of suppression allowed). 

\paragraph{Performance measurement} In the final step, the evaluation is conducted by comparing four performance measures, \ie\ classification accuracy~($Acc$), precision~($Prec$), recall~($Rec$), and F\textsubscript{1}~score, defined in terms of number of true positives~($TP$), true negatives~($TN$), false positives~($FP$) and false negatives~($FN$) as follows:
\begin{align*}
    Acc = \frac{TP+TN}{TP+TN+FP+FN}
\end{align*}
\begin{align*}
    Prec = \frac{TP}{TP+FP} \qquad Rec = \frac{TP}{TP+FN} \qquad F_1 &= 2 \frac{Prec \cdot Rec}{Prec+Rec}
\end{align*}

\begin{figure}[h]
    \centering
    \includegraphics[width=\linewidth]{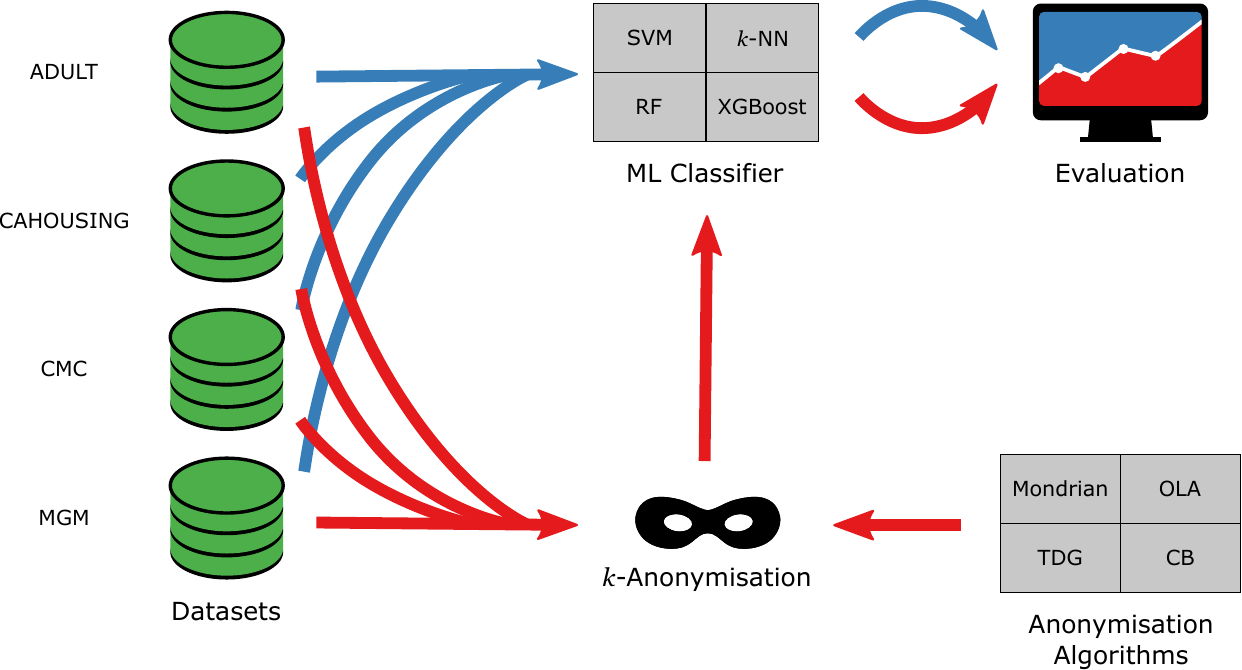}
    \caption{Overview of the experimental setup.} 
    \label{fig:setup}
\end{figure}

\autoref{fig:setup} provides an overview of our overall setup:
\begin{enumerate}
    \item Every experiment consists of a dataset (and a respective classification task) and an ML method, which is applied to the anonymised data.
    \item Each dataset is anonymised with different $k$-anonymisation algorithms using a multitude of values for~$k$ ($k=2, \ldots, 100$), resulting in multiple anonymised versions of the original dataset.
    \item The ML methods are applied to each of these newly generated datasets and the results are compared to the results obtained by each ML method on the non-anonymised data.
\end{enumerate}

For our experiments, we implemented a simple and easy-to-use environment within the software framework Python 3.7 (Python Software Foundation, USA). The datasets are stored in \texttt{csv} format and can easily be imported. Experiments can be executed directly from the command line. The configuration of the entire evaluation procedure is stored in a main experiment file. The addition of new datasets and algorithms only requires editing the main script and adding the necessary files in the folder structure. The entire code is available at \href{https://github.com/fhstp/k-AnonML}{github.com/fhstp/k-AnonML} and shall provide a common basis for extensions by other researchers.

\subsection{Datasets}
\label{sub:datasets}

In the experiments, four real-world datasets are used:

\textsc{adult}: The \emph{Adult Dataset}\footnote{\url{https://www.kaggle.com/uciml/adult-census-income}} (also known as \emph{Census Income Dataset}) contains 45,222 entries derived from the 1994 United States census database. Following the usual parameters for this use case, we use the attributes \texttt{sex}, \texttt{age}, \texttt{race}, \texttt{marital-status}, \texttt{education}, \texttt{native-country}, \texttt{workclass}, and \texttt{occupation} as QIDs and \texttt{salary-class} as the binary target variable (with two categories, \texttt{<=50K} and \texttt{>50K}). The generalisation hierarchies employed are as in \textcite{Prasser2014}. We use the predefined training and test split provided with the dataset for our experiments.

\textsc{cahousing}: The \emph{California Housing Prices Dataset}\footnote{\url{https://www.kaggle.com/camnugent/california-housing-prices}} contains 20,640 en\-tries. We choose the attributes \texttt{housing\_median\_age}, \texttt{median\_house\_value}, and \texttt{median\_income} as well as the coordinates (\texttt{longitude} and \texttt{latitude}) as QIDs (as these seem most relevant concerning privacy) and \texttt{ocean\_proximity} as the target variable. Three target values (\texttt{NEAR BAY}, \texttt{ISLAND} and \texttt{NEAR OCEAN}) were merged (giving a new, larger \texttt{NEAR OCEAN} class) to account for the imbalance in the data, leading to three classes \texttt{<1H OCEAN}, \texttt{INLAND}, and \texttt{NEAR OCEAN}. 
Data samples with missing values were removed. The generalisation hierarchies are visible in Figures~\ref{fig:cah_age} to \ref{fig:cah_value} in the supplementary material.

\textsc{cmc}: The \emph{Contraceptive Method Choice Dataset}\footnote{\url{https://archive.ics.uci.edu/ml/datasets/Contraceptive+Method+Choice}} contains 1,473 entries. The dataset is a subset of the 1987 National Indonesia Contraceptive Prevalence Survey\footnote{\url{https://microdata.worldbank.org/index.php/catalog/1398/study-description}}, which contains demographic and socioeconomic characteristics of nonpregnant women as well as the type of contraception they used. We select three attributes (\texttt{wife\_age}, \texttt{wife\_edu}, \texttt{num\_children}) as QIDs and \texttt{contraceptive\_method} as the target variable (with three possible values: \texttt{no\_use}, \texttt{short-term} and \texttt{long-term}). Our choice of QID is consistent with \textcite{Last2014}. The generalisation hierarchies are visible in Figures~\ref{fig:cmc_age} to \ref{fig:cmc_edu} in the supplementary material.

\textsc{mgm}: The \emph{Mammographic Mass Dataset}\footnote{\url{https://archive.ics.uci.edu/ml/datasets/Mammographic+Mass}}~\parencite{elter2007prediction} contains 830 entries with data from mammography analyses using the Breast Imaging-Reporting and Data System (BI-RADS), patient age and ground truth, \ie\ whether the tissue lesions are malignant or benign. We use all attributes (\texttt{age}, \texttt{shape}, \texttt{bi\_rads\_assessment}, \texttt{margin} and \texttt{density}) as QIDs and \texttt{severity} as the target variable (with two possible values: \texttt{benign} and \texttt{malignant}). Again, we employ the same QIDs as in \textcite{Last2014} to ensure comparability of results. The generalisation hierarchies are visible in Figures~\ref{fig:mgm_age} to \ref{fig:mgm_shape} in the supplementary material.

\section{Experimental Results}
\label{sec:results}

\begin{figure*}[h!]
    \centering
    \includegraphics[width=\linewidth]{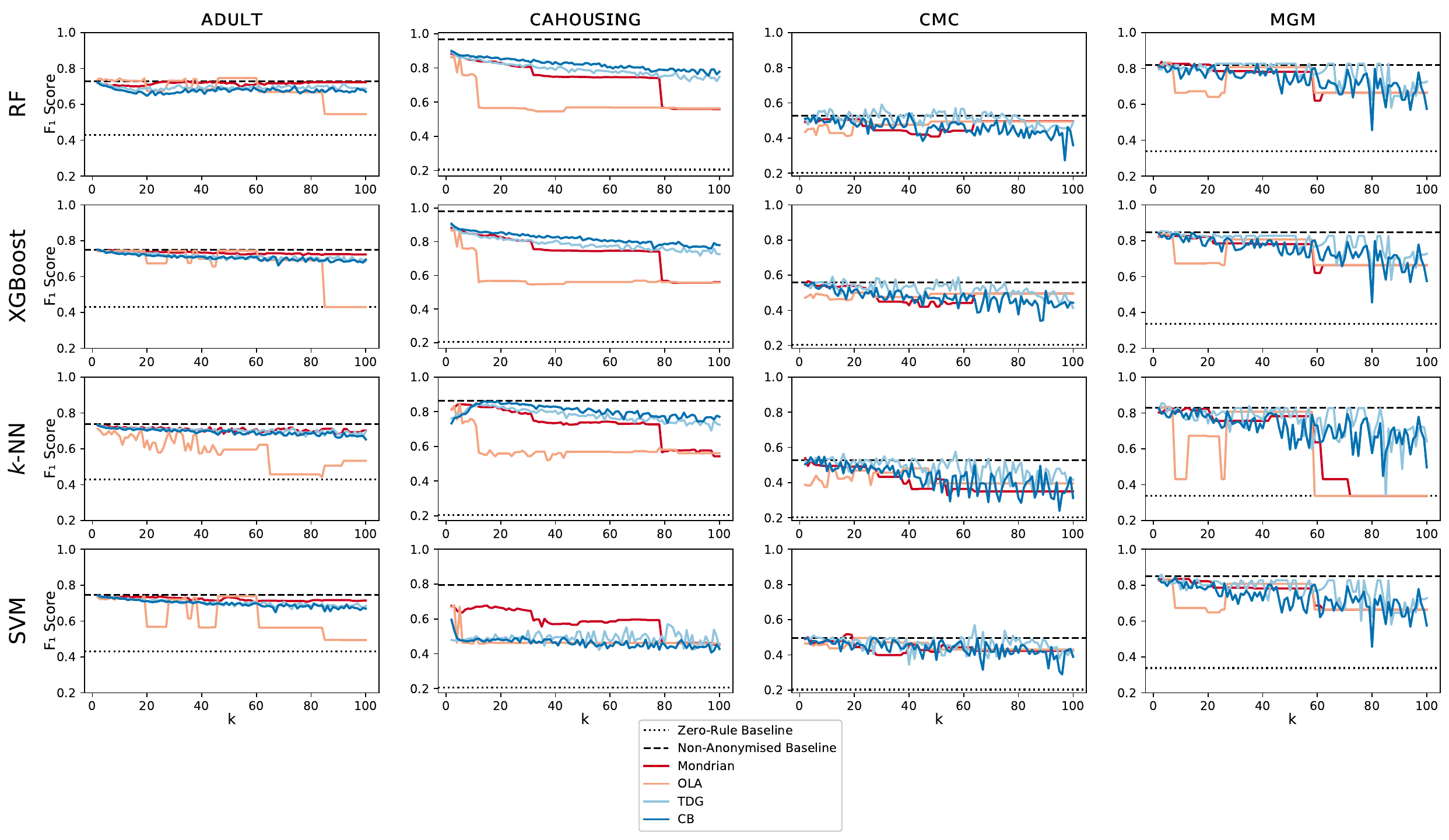}
    \caption{Overview of the performance in terms of F\textsubscript{1} scores
    for all four datasets, all four classifiers and all anonymisation methods for $k$ ranging from 2 to 100.}
    \label{fig:overview}
\end{figure*}

The classification results in terms of F\textsubscript{1}~scores are presented in \autoref{fig:overview} (for classification accuracy, precision, and recall refer to \autoref{fig:overviewAccuracy}, ~\autoref{fig:overviewPrecision} and~\autoref{fig:overviewRecall} in the supplementary material). Each row represents one of the four classifiers examined and each column corresponds to one of the four datasets. Each subfigure contains the results obtained with the four investigated anonymisation algorithms, each with ascending $k$ values ($k \in \{2, \ldots, 100\}$). The dotted lines in \autoref{fig:overview} represent the zero-rule baseline and the dashed lines represent the non-anonymised baseline, \ie\ the classification performance achievable on non-anonymised data.

In general, we can observe that OLA produces the most atypical results compared to the other algorithms. The performance curves for OLA show large jumps over different $k$ values and a highly oscillating behaviour, \eg\ for $k$-NN and SVM in combination with the \textsc{adult} dataset, and even performance values reaching the ZRB, \eg~for OLA and XGBoost with the \textsc{adult} dataset as well as Mondrian and OLA with $k$-NN and the \textsc{mgm} dataset. The results for OLA were obtained with a suppression level of 3\% and the \textit{gweight} metric. A detailed analysis of the OLA algorithm and the influence of suppression and different $k$ values on ML results is given in \autoref{sub:ola}. 

The results of Mondrian, TDG and CB show more consistent trends. Clearer differences between these three methods can be observed for the \textsc{cahousing} and \textsc{mgm} datasets. In general, the results for Mondrian seem to be less noisy (oscillating) than those obtained for TDG and CB. For more details on the behaviour of TDG and CB refer to~\autoref{subsub:tdg_cb}.

We can observe different shapes of performance curves in the experiments, with the curves either falling or remaining almost at a constant level. For some ranges of $k$, an increase in classification performance for increasing $k$ values can be observed, \eg\ for $k$-NN and the \textsc{cahousing} dataset anonymised with TDG and CB. For smaller $k$ values, these algorithms lead to much weaker results compared to Mondrian. As $k$ increases up to $k=20$, the performances of TDG and CB increase, but thereafter they decrease as expected. This is remarkable as we would expect that with increasing $k$, the classification accuracy would in general degrade. This seems not to be the case in all situations and may be a dataset-dependent effect. Since it occurs only in isolated cases, we would consider these results rather as outliers. We assume that the improvements in classification performance may occur in cases where the generalisation hierarchy generated by the anonymisation algorithm correlates well with the class structure in the dataset and thus helps the classifier to partition the classes more easily.

Nearly constant curves can be observed for Mondrian, TDG and CB with the \textsc{adult} dataset. For all other scenarios, decreasing curves can be observed (which is generally the expected behaviour as $k$ increases). In the following sections we discuss and analyse in more detail our observations for the different experiment configurations.

\subsection{Observations on Individual Datasets}

In the following, we describe the observations made in our experiments for the individual datasets (across classifiers and anonymisation methods) and identify common patterns. We consider other perspectives (\eg\ for individual classifiers) further below.

The results for the \textsc{adult} dataset are consistent among all classifiers. All anonymisation algorithms, with the exception of OLA, behave similarly -- as $k$ increases, their performance decreases slightly. Mondrian exhibits the most robust behaviour, especially at larger values of $k$, where the best performances are achieved. The only outlier in anonymisation methods is OLA, which shows strong fluctuations in the resulting classification performances consistently across all classifiers. The performance curves are relatively volatile for $k$ values below $46$. For $k$-NN and SVM, a significant drop occurs at $k=65$ and $k=61$, respectively, whereas for the tree-based classifiers this drop emerges much later at $k=85$ (for XGBoost even reaching the ZRB). The underlying cause of this significant decrease in performance is presented in \autoref{sub:ola}. For the tree-based classifiers, the volatility in the performance curve (up to $k=46$) is much smaller than for $k$-NN and SVM; moreover, the results obtained with the tree-based classifiers are more accurate and comparable with other anonymisation algorithms (excluding the previously mentioned results above $k=85$).

For the \textsc{cahousing} dataset, Mondrian shows consistent results across all classifiers, with a moderate decrease up to $k=79$, followed by an abrupt drop of up to 17\% (see \autoref{sub:mondrian} for a more detailed analysis). The performance of the OLA algorithm shows a similar behaviour, but the drop already occurs for much smaller $k$ (for RF, $k$-NN and XGBoost at $k=12$ and for SVM at $k=6$). Therefore, OLA generally performs very badly for this dataset. For tree-based ML models, the results with TDG and CB are robust as $k$ increases. They decrease almost monotonously without large or abrupt changes. The results for SVM are generally much worse than those of the other classifiers; moreover, TDG and CB perform much worse than Mondrian (up to $k=80$, after which they perform similarly). 

For \textsc{cmc}, the individual performance curves obtained with the different anonymisation algorithms have similar progressions across the classifiers. Similarly to the other datasets, TDG and CB show more fluctuations with varying $k$ than OLA and Mondrian. The results obtained with the OLA-anonymised data deviate less from the results of the other algorithms, with the exception of smaller $k$ values, for which the loss of information introduced by OLA seems to be slightly higher than for the other anonymisation algorithms with the same $k$.

The results for the \textsc{mgm} dataset show a relatively consistent performance across all classifiers. The anonymisation algorithms all start with very high F\textsubscript{1}~score values, with Mondrian, TDG and CB moderately decreasing in performance up to $k=59$, after which Mondrian exhibits a severe performance loss (even reaching the ZRB for $k$-NN). For OLA, we observe a rapid drop of 18\%--47\% (depending on the classifier) between the $k$ values 8 to 26. Thereafter, OLA performs comparably to the other algorithms up to $k=59$. From $k=59$, Mondrian and OLA have a 18\%--58\% performance loss. Classification results obtained for TDG and CB show strongly increased fluctuation from $k=59$ upwards. Detailed analyses of these behaviours are presented in \autoref{sub:ola}, \autoref{sub:mondrian} and \autoref{subsub:tdg_cb}.

\subsection{Observations on Individual Classifiers}

The comparison of the classifiers with respect to performance reveals that the tree-based classifiers, \ie\ RF and XGBoost, perform considerably better over the different datasets. In some experiments, this is already clearly evident from the absolute F\textsubscript{1}~score, in others from the robustness of the results across the different $k$ values. The better performance of RF and XGBoost is particularly evident for the \textsc{cahousing} dataset in both settings, \ie\ both using non-anonymised and anonymised data. One explanation could be that the two tree-based classifiers handle the generalisation (\ie\ successively removing information by eliminating significant digits) of the two QIDs \texttt{longitude} and \texttt{latitude} more efficiently. For the \textsc{cahousing} dataset, the linear SVM fails completely for the anonymised data (even the curve for Mondrian is clearly inferior to the other classifiers) and $k$-NN performs significantly worse compared to the tree-based classifiers especially for smaller $k$ values.

For the \textsc{mgm} dataset, $k$-NN exhibits strong performance variations, which are most evident for OLA and Mondrian. However, for CB and TDG the fluctuations also increase considerably for larger $k$ values. One explanation for the weak performance of $k$-NN in various experiments may be the choice of the hyperparameter $k$ (\ie~the number of neighbours), since the algorithm is very sensitive to this data-dependent parameter. 

\subsection{Robustness Considerations for OLA}
\label{sub:ola}

In our experiments, we observe strong performance variations (fluctuations as well as abrupt changes) in ML experiments when OLA is employed for anonymisation. OLA seems to be highly dependent on the choice of generalisation hierarchies for the QIDs of a given dataset and the synergy (or lack thereof) of the hierarchy with the chosen information loss metric. When the VGHs and metric are not well aligned, results can be highly erratic and exhibit a large degree of seemingly random fluctuation, due to OLA frequently switching the chosen node in the VGH lattice. This is particularly evident for the \textsc{adult} and \textsc{mgm} datasets, where results seem especially unstable, while the \textsc{cahousing} and \textsc{cmc} datasets show less of this behaviour, \ie\ a more monotonous progression with increasing $k$. Using the \textsc{adult} dataset as an example, we examine this behaviour and the effects of the suppression level and four metrics commonly used for OLA: precision (\textit{prec})~\parencite{sweeney2002achieving}, generalisation weight (\textit{gweight})~\parencite{samarati2001protecting}, average equivalence class size (\textit{aecs})~\parencite{LeFevre2006} and discernability metric (\textit{dm})~\parencite{BayardoAgrawal2005}.

\begin{figure*}[h]
    \centering
    \includegraphics[width=\linewidth]{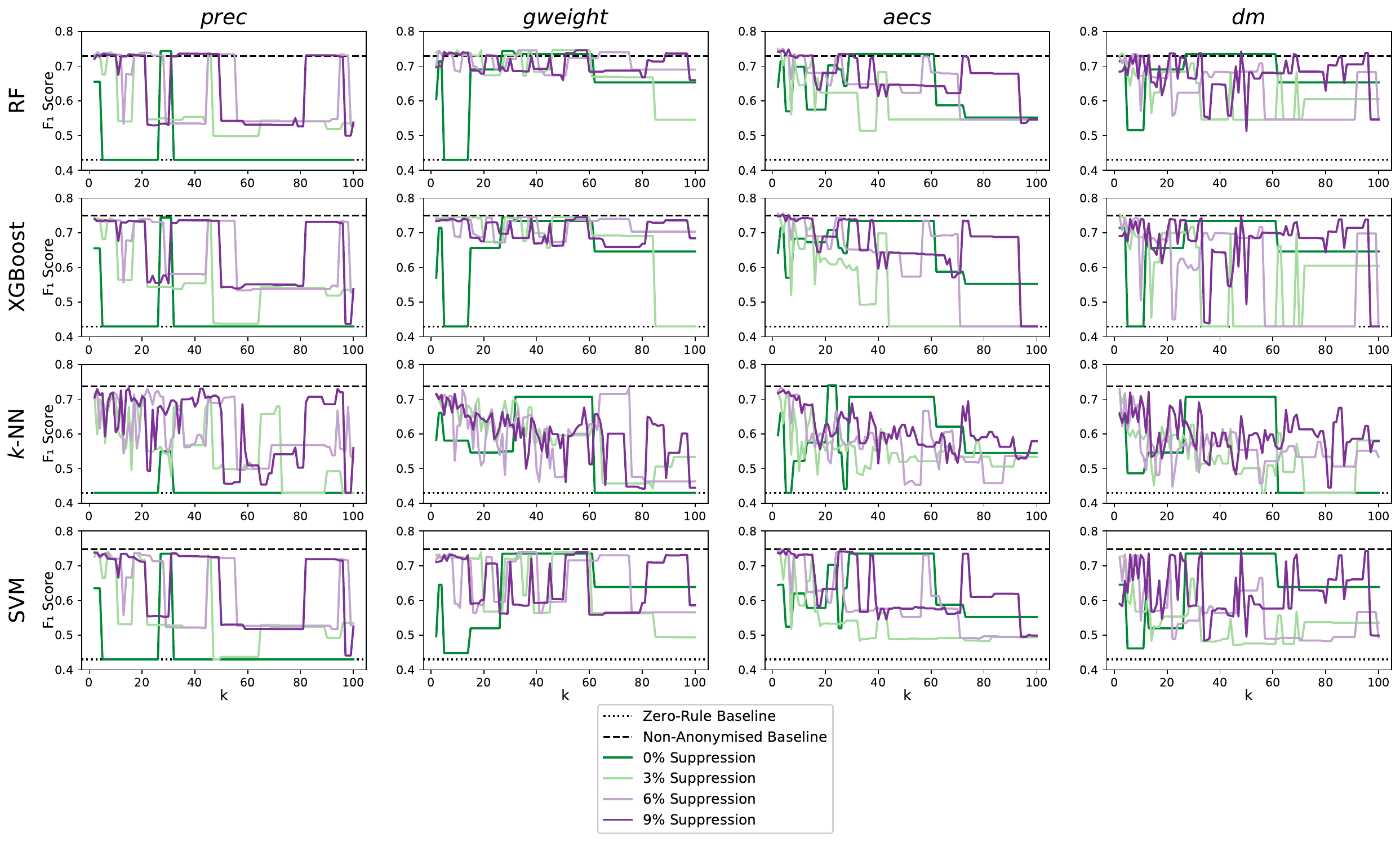}
    \caption{Overview of more detailed classification results for OLA applied on the \textsc{adult} dataset to further investigate the unstable classification results. Results are shown for different metrics (\textit{prec}, \textit{gweight}, \textit{aecs} and \textit{dm}) and different suppression levels: 0\% (no suppression), 3\%, 6\% and 9\%.}
    \label{fig:adultOLA}
\end{figure*}

As \autoref{fig:adultOLA} shows, the investigated behaviour does not strictly depend on the choice of the suppression level. Our results indicate that allowing higher percentages of suppressed entries gives OLA more freedom to switch nodes more often. Allowing \emph{any} suppression (as opposed to none) seems to give rise to an increase in fluctuation. For some combinations of metric and classifier (\eg\ \textit{dm} and RF), allowing up to 9\% suppression causes an evident increase in fluctuation compared to the scenario with only up to 3\% suppression. Note that this does not necessarily mean that ``more suppression'' and ``more fluctuation'' correlate in general.

Our experiments indicate that the choice of metric is important and can have a significant effect on classification results. However, the effect of the employed metric and the identification of an optimal metric largely depend on the specific dataset and use case.
Thus, we cannot derive a general recommendation for a specific metric. Experimental results obtained on the \textsc{adult} dataset (see \autoref{fig:adultOLA}) show a clear benefit in using \textit{gweight} across the classifiers RF, XGBoost and SVM. The metric \textit{aecs} performs better in conjunction with $k$-NN.

\begin{figure*}[h]
    \centering
    \includegraphics[width=\linewidth]{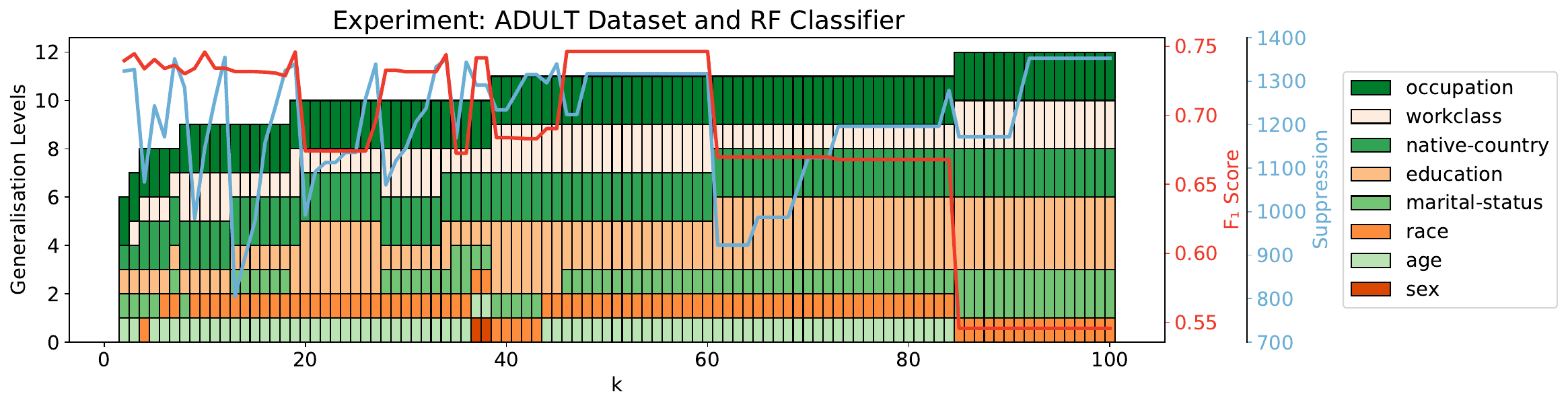}
    \caption{Analysis of the combined effects of generalisation and suppression for OLA (with the \textit{gweight} metric), RF as classifier and the \textsc{adult} dataset. Each coloured bar corresponds to a feature (\eg\ \texttt{occupation}) and its height to the generalisation level (in the generalisation hierarchy). The blue curve represents the number of suppressed records and the red curve the F\textsubscript{1}~score obtained for various $k$ values.} 
    \label{fig:generalisationandsuppression}
\end{figure*}

Finally, we investigate in more detail the individual effects of generalisation and suppression as well as their combined effects on classification performance. For this investigation, we consider the example of OLA with the \textit{gweight} metric, applied to the \textsc{adult} dataset with a suppression of up to 3\%.

To better understand the investigated dataset, we calculate Spearman's rank correlation coefficients between the target variable and the features (categorical features were one-hot encoded). The correlation matrix is presented in the supplementary material (\autoref{fig:corr_matrix_adult}). A moderately positive correlation~($r_s = 0.45$) exists for the category \texttt{married-civ-spouse} of the \texttt{marital-status} feature. Moreover, for the features \texttt{sex}~($r_s = 0.22$) and \texttt{age}~($r_s = 0.27$), the category \texttt{exec-managerial}~($r_s = 0.21$) of the \texttt{occupation} feature and the category \texttt{never-married}~($r_s = -0.32$) of the \texttt{marital-status} feature a weak correlation can be observed.

\autoref{fig:generalisationandsuppression} shows the generalisation levels for the individual features in the dataset for $k$ values ranging from 2 to 100. The red line shows the achieved F\textsubscript{1}~score and the blue line the suppression level (number of records suppressed). The highest drop in the F\textsubscript{1}~score coincides with the complete generalisation of the \texttt{marital-status} feature at $k=85$. This is in accordance with the correlation analysis (moderate correlation with the target variable) showing that \texttt{marital-status} is essential for classification. We further observe an effect of the generalisation level of the \texttt{education} feature; the complete generalisation of this feature (in the ranges starting at $k=20$, $k=39$ and $k=61$) causes a significant drop in performance. Even though the correlation of the \texttt{education} feature and the target variable is negligible (the highest correlation exists for the categories \texttt{bachelors} ($r_s = 0.18$) and \texttt{masters} ($r_s = 0.17$)), the model can still exploit the information in this feature.

While our results show increasing F\textsubscript{1}~score fluctuation for OLA when suppression is applied (as depicted in \autoref{fig:adultOLA} for RF and \textit{gweight}), the number of suppressed records, however, does not directly influence the F\textsubscript{1}~score. This is particularly noticeable between $k=62$ and $k=84$ in \autoref{fig:generalisationandsuppression}, where the number of suppressed records increases while the F\textsubscript{1}~score remains constant.
The reason for this is the node selection process of OLA: Increasing the number of suppressed records allows OLA to keep using a node with a lower generalisation level for a longer run of subsequent $k$. 

In contrast, OLA without suppression applies stronger generalisation from the beginning and, therefore, selects nodes more consistently over consecutive $k$ values (with fewer node changes). This behaviour results in less fluctuation of the F\textsubscript{1}~score, as for instance apparent in \autoref{fig:adultOLA} for RF and \textit{gweight}.
While the use of suppression leads to stronger fluctuation of the OLA results, the magnitude of the individual drops is lower on average when compared to OLA without suppression. Overall, we cannot settle the question whether suppression is good or bad in general; suppression and generalisation complement each other and their joint effect depends on the precise nature of their interaction in the specific use case.

\subsection{Robustness Considerations for Mondrian}
\label{sub:mondrian}

\autoref{fig:overview} shows that the F\textsubscript{1}~scores obtained in the classification experiments with Mondrian as the anonymisation algorithm remain fairly stable for most values of $k$, \ie\ almost no fluctuations and in most cases only a slight performance loss with increasing $k$. There is, however, also some deviating behaviour of F\textsubscript{1}~score progression where the score abruptly drops, which deserves a closer investigation. In order to find the reason for these drops, we further analysed the results of Mondrian for the two datasets \textsc{cahousing} and \textsc{mgm} where these abrupt performance drops mostly occur.

\begin{figure*}[h]
    \centering
    \includegraphics[width=\linewidth]{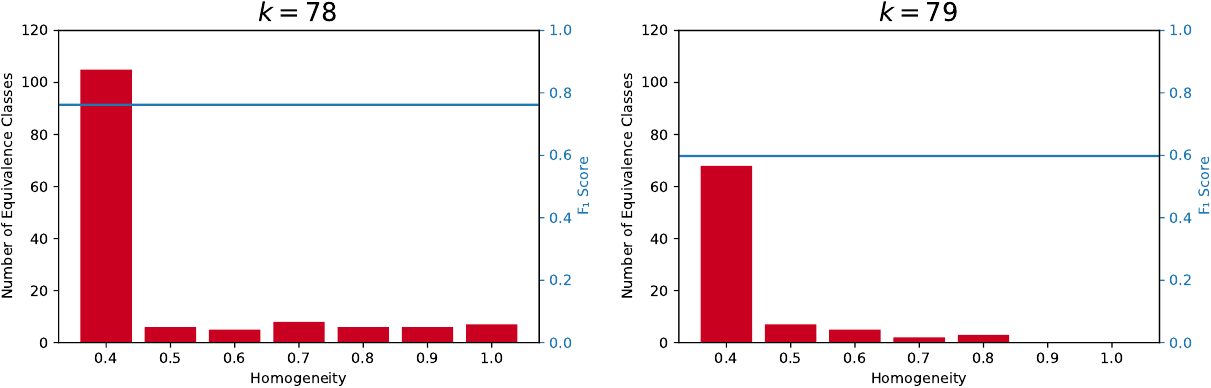}
    \caption{Number of equivalence classes, their respective homogeneity and the F\textsubscript{1} score for the RF classifier on the \textsc{cahousing} dataset. For $k=78$, the number of equivalence classes with low homogeneity is large (above 100). For $k=79$, this number abruptly decreases to 68.}
    \label{fig:cahousinghisto}
\end{figure*}

The biggest drop of the F\textsubscript{1}~score for the \textsc{cahousing} dataset can be observed from $k=78$ to $k=79$ (across all classifiers). This seems to be caused by a rather large information loss due to switching from the penultimate generalisation step to the complete generalisation of the \texttt{latitude} feature for every equivalence class. As a consequence, the number of equivalence classes is reduced from $143$ to $85$ and the homogeneity within each equivalence class decreases, as depicted in \autoref{fig:cahousinghisto}. ``Homogeneity'' here refers to the distribution of predicted classes in the equivalence classes: Higher homogeneity means that the equivalence classes contain mostly samples of one predicted class each, which makes it easier for the classifier to model the data; low homogeneity means that many samples from different predicted classes share the same equivalence class, making classification difficult (or even impossible) and thus leading to a decrease in overall performance.

\begin{figure*}[!ht]
    \centering
    \includegraphics[width=\linewidth]{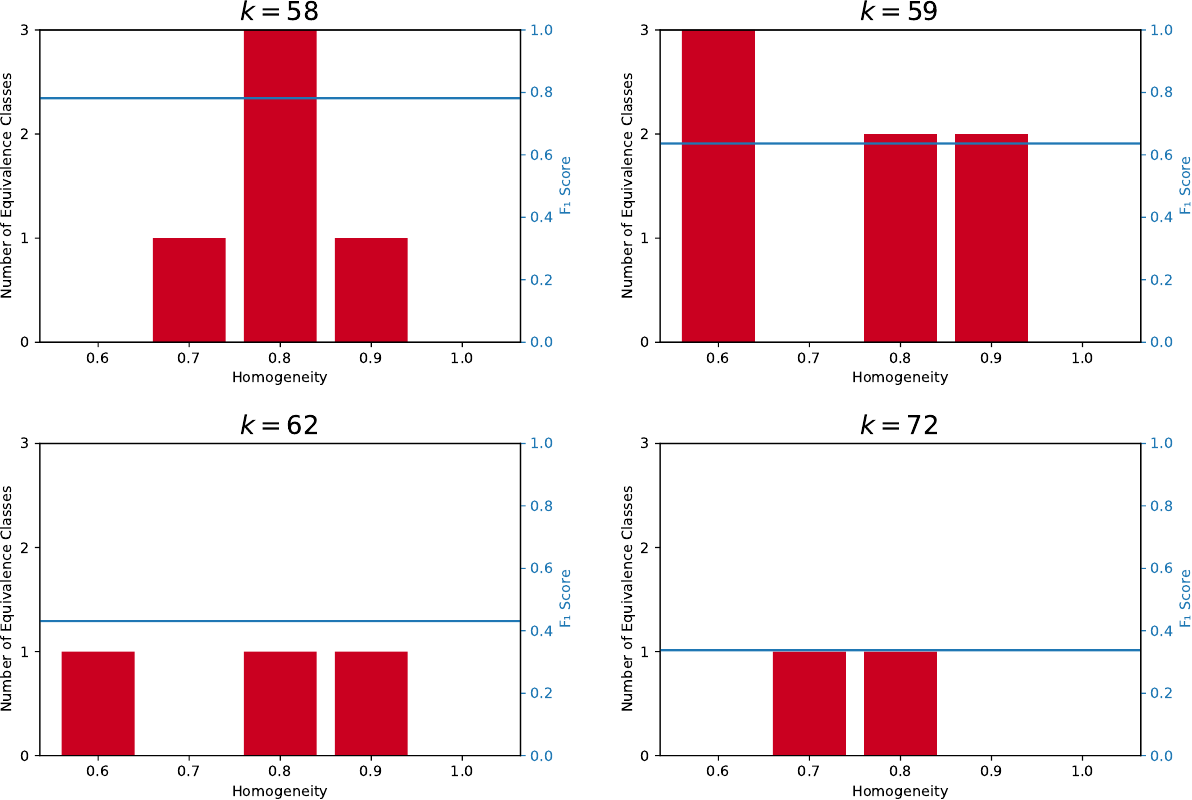}
    \caption{Number of equivalence classes, their respective homogeneity and the F\textsubscript{1} score of the $k$-NN classifier on the \textsc{mgm} dataset. The transition from $k=58$ to $k=59$ shows a performance loss caused by less homogeneous equivalence classes. The subsequent transitions to $k=62$ and $k=72$ show a performance loss primarily caused by the reduction of the number of equivalence classes.}
    \label{fig:mgmhisto}
\end{figure*}

For the \textsc{mgm} dataset, the worst declines in classification performance occur at the step from $k=58$ to $k=59$ as well as at $k=62$ and $k=72$. The number of equivalence classes and their respective homogeneity for each of those $k$ values is shown in \autoref{fig:mgmhisto}.

The iteration $k=58$ results in $5$ distinct equivalence classes and a fairly high homogeneity for each of the equivalence classes. The number of equivalence classes for the subsequent iteration $k=59$ increases to $7$, while the homogeneity of the equivalence classes decreases (the bars shift to the left). This is caused by the weaker generalisation of the \texttt{age} attribute, while at the same time losing all information for the \texttt{shape} attribute due to its complete generalisation. The reason for the drop at $k=62$ lies in the reduction of equivalence classes from $7$ to only $3$, which results in a generalisation of the equivalence classes that differs only in the \texttt{age} attribute. Iteration $k=72$ shows a similar result, with only $2$ distinct equivalence classes remaining. This especially degrades the performance of $k$-NN, which has a very local view and not enough flexibility to compensate for this strong data reduction. The other classifiers are not affected by this generalisation step.

Overall, we can conclude that stronger generalisation can degrade classification performance but does not necessarily have to. The actual loss in classification performance depends on (at least) the modelling capabilities of the classifier and the correlation of the generalised features with the target variable for the classification task: The stronger the correlation with the target variable, the higher the chance that generalisation will degrade the classification performance.

\subsection{Strong Fluctuations for TDG and CB}
\label{subsub:tdg_cb}

As depicted in \autoref{fig:overview}, the calculated F\textsubscript{1}~scores of the anonymisation algorithms TDG and CB fluctuate strongly. This result is quite contrary to the scores of Mondrian, for which the values in most configurations are more stable and the performance degrades monotonically with increasing $k$.

Since TDG and CB utilise a random function in order to pick random initial data records (as described in \autoref{subsub:tdg} as well as \autoref{subsub:cb}), we examined whether the observed fluctuation occurs due to this randomness in the algorithms. To that end, we recomputed the results for these algorithms with fixed random seeds. Although the F\textsubscript{1} score results for the individual iterations of $k$ differ from the previous results, the fluctuations are similarly strong in terms of magnitude and frequency. Therefore, we conclude that the random factor has either no or only minor influence on the fluctuation.

\begin{figure*}[!ht]
    \centering
    \includegraphics[width=0.6\linewidth]{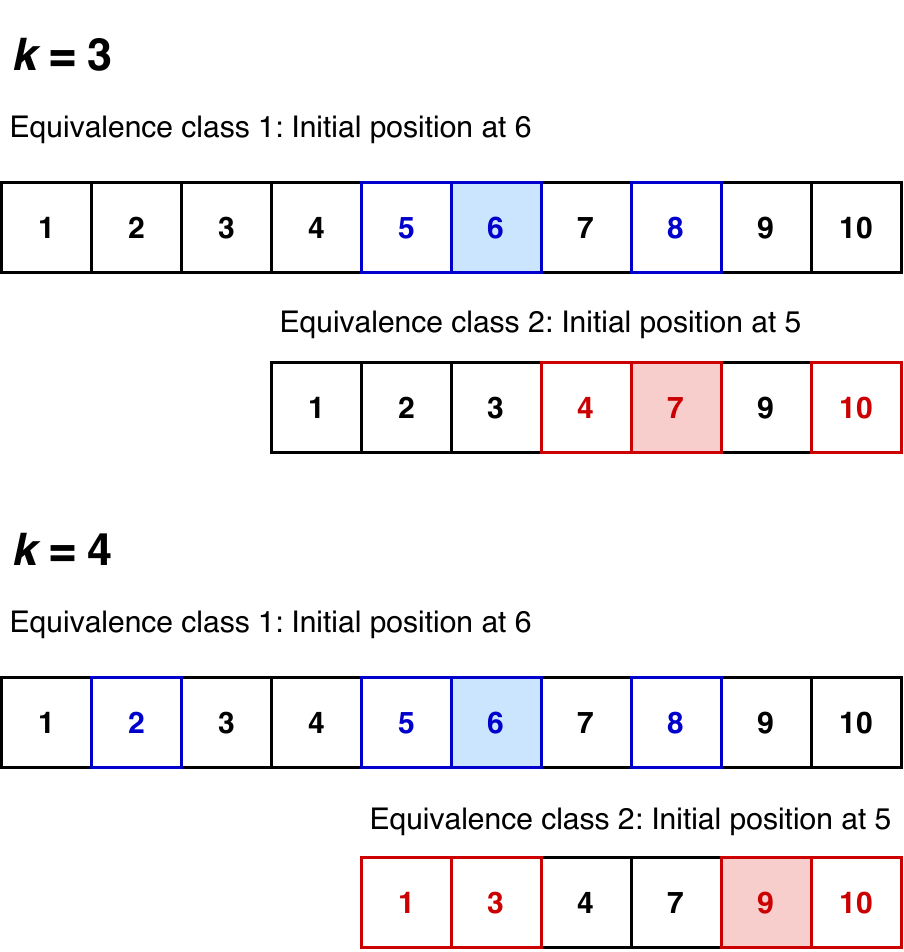}
    \caption{Abstract functioning of equivalence class building of TDG and CB for $k=3$ and $k=4$ utilising a fixed random seed. The algorithm chooses an initial data record for an equivalence class, illustrated as a box with shaded background, at a specific position yielded by a random function and puts other records into the equivalence class by algorithm-specific means in order to fulfil the $k$-anonymity requirement. Although the positions yielded by the random function remain the same for different iterations of $k$, the data record at this position can be different. Since for $k=4$, the record at position $2$ is already put into equivalence class $1$ (in contrast to the iteration $k=3$, where the equivalence class is just $\{5,6,8\}$), the initial record at position $5$ for equivalence class $2$ is different for the two iterations, resulting in vastly different equivalence classes overall.}
    \label{fig:tdg_cb}
\end{figure*}

Further analysis showed that the fluctuation might result from the nature of the algorithms. Both TDG and CB (while working differently in detail) generate equivalence classes based on just the data records, while the other anonymisation algorithms in our study generate equivalence classes based on additional (structural) information such as the domain space of attributes and assign data records to the equivalence classes accordingly. For both TDG and CB, the minimum size of the equivalence classes increases with $k$ and data records of equivalence classes with size less than $k$ are merged with other equivalence classes. Since the size of the equivalence class increases with increasing $k$, the initial data records chosen for each equivalence class are different for each choice of $k$ (even when the algorithm utilises a fixed random seed). \autoref{fig:tdg_cb} illustrates an example for the iterations $k=3$ and $k=4$ as well as a dataset size of $10$. In this example, the CB algorithm for $k=3$ may choose the data record at position $5$ of the remaining dataset (without the records inside the first equivalence class), which was retrieved by the random function with a fixed seed, as the starting point for building the second equivalence class and assign the $2$ nearest records to that equivalence class in order to fulfil the $k$-anonymity requirement. In contrast, for the iteration $k=4$ the algorithm might place an additional data record into the first equivalence class, and this record could have been at a position lower than the record chosen for the second equivalence class during the iteration $k=3$. Therefore, the data record at position $5$ is now a different one compared to the previous iteration. This other record is now used as the starting point for the second equivalence class, and the nearest members for this record most likely differ as well, resulting in a potentially vastly different equivalence class. Therefore, the members of the equivalence classes may change for each iteration \emph{despite} choosing a fixed seed. As a result, the generalised equivalence classes can change substantially even when comparing consecutive iterations of $k$. 

In addition to the fluctuation patterns discussed above, we observe that most strong drops of the F\textsubscript{1}~score occur in cases where a large proportion of equivalence classes are subject to the same generalisation, especially in case of the strongest possible generalisation. For instance, such a drop is evident in \autoref{fig:overview} from $k=84$ to $k=85$ for TDG in conjunction with the $k$-NN classifier applied to the \textsc{mgm} dataset.

\subsection{Importance of Dataset Preparation}
\label{sub:datasetprep}

\begin{figure*}[h!]
    \centering
    \includegraphics[width=0.7\linewidth]{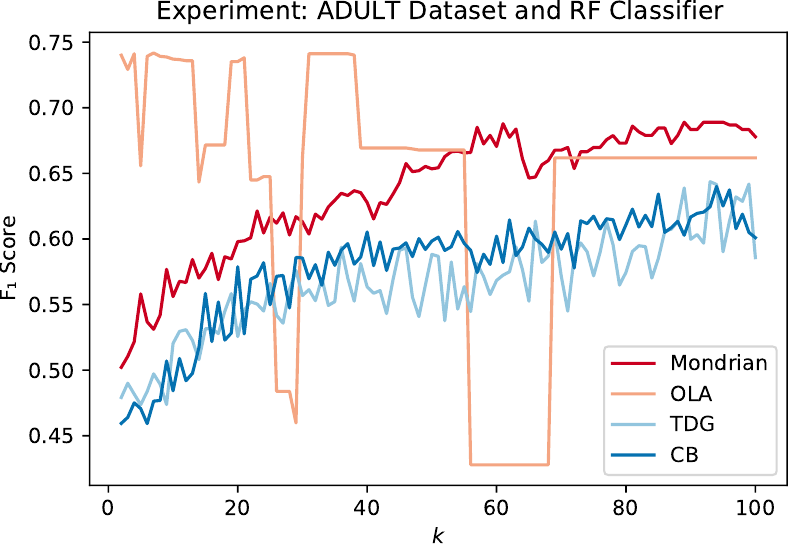}
    \caption{Example of classification results for RF on a randomly selected validation subset (of the training set) of the \textsc{adult} dataset where a naive one-hot encoding is used for numerical attributes that are generalised, leading to suboptimal results.}
    \label{fig:preprep}
\end{figure*}

Proper data anonymisation requires adequate preparation of the data. The related data transforms can have strong effects on the modelling abilities of the classifiers and the classification performance. In the following, we present the insights gained regarding data preparation. During data preparation, categorical attributes (\eg\ \texttt{sex} or \texttt{native-country}) are commonly converted into sparse binary numerical arrays using one-hot encoding. Numerical values (\eg\ \texttt{age}) are usually generalised by replacing them with increasingly wide intervals. In our first experiments, we overlooked the fact that such intervals were not directly interpretable as numerical values themselves, since they were naively represented as strings (\eg\ ``[1-5]'') and then converted using one-hot encoding. Thus, the generalisation of even just one data record was sufficient to lead to one-hot encoding the whole attribute. This resulted in an interesting effect during classification: We obtained low classification performance for small $k$, which improved with increasing $k$ (see \autoref{fig:preprep}). We assume that this counter-intuitive behaviour stems from the extraordinarily high dimension of the one-hot encoded numerical attributes (curse of dimensionality) preventing robust modelling by the classifiers, as well as the fact that the classifiers could not make use of the numeric nature of the attributes in question. With increasing $k$, more and more numerical values were generalised, leading to a reduced number of different intervals, which improved the modelling capabilities of the classifiers. Learning from these preliminary experiments, we decided to replace the one-hot encoded interval strings (\ie\ categorical data) in the anonymised data with their averages (\ie\ numerical data) before providing the data to the ML classifiers. Thereby, the dimensionality is reduced and the numeric nature of the attributes preserved, allowing the algorithms to appropriately model the data.

Overall, this serves to point to the (somewhat obvious) fact that preserving the numericality of numeric attributes is key to effective classification. While this might indicate that using microaggregation, which is by design more suited for numeric values~\parencite{domingo2005ordinal}, should generally be preferred for numeric attributes, see the next section for our experimental findings on this issue.

\subsection{Comparison of Generalisation and Suppression with Microaggregation}

A comparison of our results with those of the systematic study conducted by \textcite{rodriguez2018does} may provide insight into whether the two families of anonymisation algorithms exhibit similar behaviour in terms of ML performance. However, a direct comparison is impossible -- neither are their algorithms available nor do they give sufficient information in their paper regarding the chosen hyperparameters for the utilised ML models. To allow for at least a partial comparison, we performed experiments with our investigated generalisation and suppression algorithms on the \textsc{adult} dataset with the same selection of QIDs they used, \ie\ \texttt{sex}, \texttt{age}, \texttt{marital-status}, \texttt{education-num}, \texttt{capital-gain} and \texttt{hours-per-week}. Please note that a direct comparison of performance is not possible, due to different implementations of the utilised bagging classifier as well as different initialisation and hyperparameters of the model. However, we can contrast performance trends.

To additionally provide a direct comparison between a microaggregation method and our results, we use the publicly available library  $\mu$-ANT\footnote{\url{https://github.com/CrisesUrv/microaggregation-based_anonymization_tool}}~\parencite{sanchez2020mu}. Similar to the metholodgy of \textcite{rodriguez2018does}, this library provides a microaggregation method using a variant of the Maximum Distance to Average Vector (MDAV) algorithm~\parencite{hundepool2003argus}, a standard microaggregation algorithm. This algorithm first creates clusters with at least $k$ similar records; then values of QIDs are replaced with cluster averages to achieve $k$-anonymity. For numerical values this approach is straightforward, but for categorical values it is more complicated. The $\mu$-ANT library implements an additional similarity measure based on the semantic meaning of categorical QIDs. The semantic information is defined using the Web Ontology Language (OWL). For our experiments, we use the OWL ontologies provided in the library for the \textsc{adult} dataset. In contrast to this approach, \textcite{rodriguez2018does} numerised the categorical QIDs before anonymisation. As a sanity check, we also numerised the categorical QIDs using $\mu$-ANT and obtained relatively similar results to the ontology-based approach.

\begin{figure*}[h!]
    \centering
    \includegraphics[width=\linewidth]{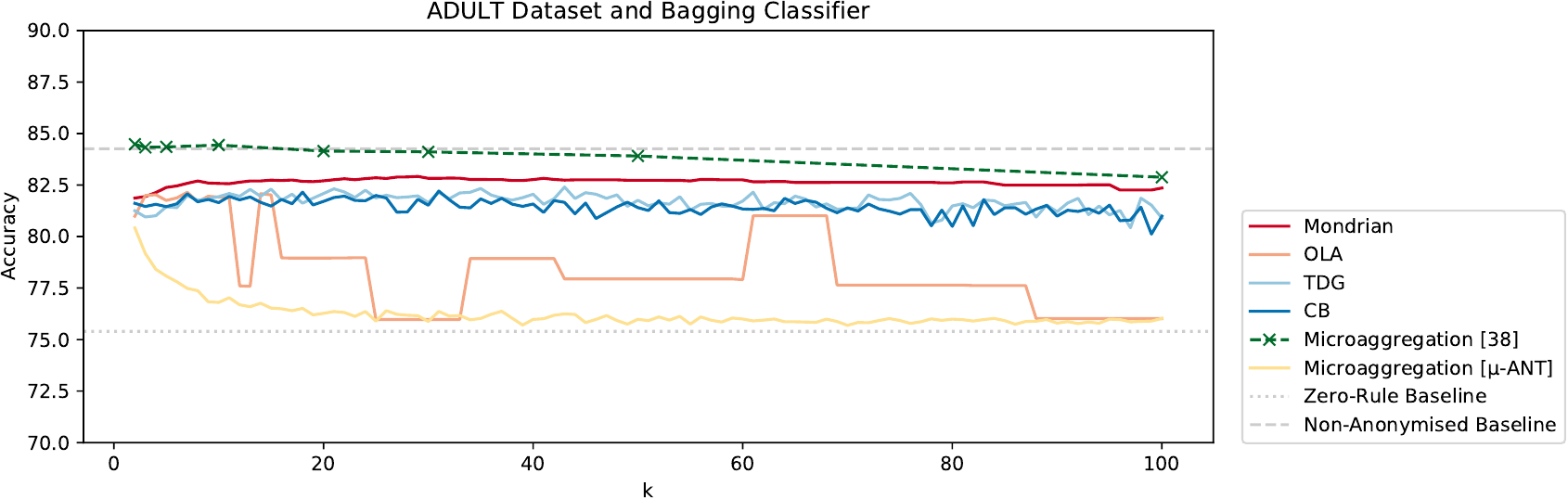}
    \caption{Comparison of the classification performance on the \textsc{adult} dataset between generalisation and suppression algorithms (\ie\ Mondrian, OLA, TDG, and CB), the microaggregation method from $\mu$-ANT (yellow line) and microaggregation results reported by \textcite{rodriguez2018does} (green line; the markers indicate the values evaluated and explicitly reported by~\textcite{rodriguez2018does}, while the curve is interpolated). The results are presented in terms of classification accuracy.}
    \label{fig:adult_micro}
\end{figure*}

\autoref{fig:adult_micro} shows (1) the results of the generalisation and suppression algorithms, \ie\ Mondrian, OLA, TDG and CB, (2) the microaggregation results obtained with $\mu$-ANT (yellow line) and (3) the microaggregation results reported in the study of \textcite{rodriguez2018does} (green markers). Note that \textcite{rodriguez2018does} do not provide sufficient information about the settings and parameters of the ML experiments performed. Consequently, while we have done our best to replicate their settings to compute the results presented in \autoref{fig:adult_micro}, the results of the green curve are nevertheless not directly comparable with the other curves.

The baseline for the non-anonymised data in the study of \textcite{rodriguez2018does} (84.63\%) and our configuration (84.26\%) are relatively similar. The generalisation methods show similar behaviour to the results in \autoref{fig:overview}, with Mondrian performing most robustly, TDG and CB showing fairly volatile behaviour but generally maintaining their performance, and OLA showing large fluctuations in the resulting classification performance. For smaller $k$, Mondrian shows similar performance values as the other generalisation algorithms and increases slightly up to $k=29$, after which it remains relatively constant. Such behaviour is not observed in the results of \textcite{rodriguez2018does}, where the performance starts relatively close to the non-anonymised baseline and decreases steadily. Comparing Mondrian and $\mu$-ANT (with the same configuration), we see two completely contrasting behaviours, with $\mu$-ANT showing a slightly worse result at $k=2$ and then exhibiting a rapid decline over the first 20 values of $k$ to finally settle near the zero-rule baseline.

Our results show inconsistent behaviour for the two microaggregation approaches. This motivates the development of a test bed in which all anonymisation algorithms and techniques (\ie\ generalisation and suppression as well as microaggregation and others) can be studied in detail using the exact same settings.

\subsection{Remarks on the Interplay of Anonymisation and Machine Learning}

We finally want to make some concluding remarks on the interplay between anonymisation and ML techniques that originate from our study. We have shown that anonymisation and ML techniques are highly dependent on each other. Both try to solve very different and partly contradictory goals, complicating their combination. Anonymisation techniques may remove information from a dataset which is important for solving a classification task and may thus degrade the classification performance. Different classifiers show different degrees of sensitive behaviour to such information loss introduced by \eg\ generalisation or suppression of values. Consequently, to achieve a good overall performance, both processes need to be optimised jointly. A central question in this context is thus how to achieve an appropriate degree of anonymisation (\eg\ a certain level of $k$-anonymity) and at the same time minimise the information loss for the given classification task. 

Our study has revealed certain patterns in the behaviour of anonymisation techniques which may be suitable starting points for further improvements to foster the compatibility of anonymisation and classification techniques. One such observed pattern is the strong fluctuation of classification performance obtained on data anonymised by CB and TDG for different $k$. Such behaviour is not desirable from the perspective of ML, because it means that the overall performance strongly depends on the selection of one particular system parameter, in this case $k$. A high sensitivity for one system parameter is in general undesired since it impedes the reliability of the overall system performance as well as the targeted optimisation thereof.
One potential improvement to mitigate this issue would be to define the initial records of all equivalence classes before assigning other records to the equivalence classes. With this change, the equivalence class creation would happen at the start of the anonymisation process. In practice, increasing $k$ by $1$ would thus mean that all initial values remain the same and a new initial value is added for the new equivalence class. The records in an equivalence class would still remain different for consecutive $k$; however, the distance of the records to the initial records should vary less.
This would reduce the fluctuations in obtained classification performance and lead to more stable results.

A second observation from our study is that the selection of features for generalisation plays a critical role for the classification performance that is achievable. The generalisation of features which are strongly correlated to the target variable of the classification task can strongly degrade the performance. For many datasets there are different ways (\ie\ different generalisation steps) to achieve the same degree of $k$-anonymity. In such situations, the generalisation that changes the strongly correlated features least should be chosen. This argument suggests a combined optimisation of anonymisation and ML methods (\ie\ choosing the optimal anonymisaton algorithm based on ML task performance instead of abstract metrics), which to us seems like a particularly promising future research direction.

\section{Conclusion}

We have presented an in-depth study into the effects of data anonymisation on classification performance. A special focus was put on anonymisation methods which build upon the principles of generalisation and suppression, since this represented a gap in the literature so far. For our evaluation, we have selected a set of four popular anonymisation techniques using generalisation and suppression as well as four heterogeneous classification methods. To reduce the bias and influence of dataset choice on our study, we have selected four different datasets on which all experiments were performed.

We compared the achievable classification performance on top of the differently anonymised data and investigated the individual behaviour of the anonymisation methods (\eg\ abrupt performance degradation and fluctuations). We investigated the internal workings of the anonymisation techniques to explain our observations.

Our results show that -- as we would expect -- with an increasingly strong $k$-anonymity constraint, the classification performance generally degrades. The amount of degradation is, however, strongly dependent on dataset and anonymisation method. Furthermore, we show that some anonymisation strategies provide a better basis for downstream classification than others. While TDG and CB show strongly varying performance for different $k$ (which originates from the heuristic they use to build equivalence classes), making it difficult to estimate the achievable classification performance in general, Mondrian shows more robust behaviour. Furthermore, in most classification experiments data anonymised by Mondrian outperforms that obtained by OLA (or provides at least the same level of performance). Thus, Mondrian can be considered the method with the most appealing properties for subsequent classification experiments. Moreover, our investigation of OLA shows that many typical data precision metrics (in particular those based on counting generalisation levels) can be misleading when trying to estimate the actual impact of anonymisation on the quality of the anonymised data, \eg\ a doubling of the measured data loss in the metric does not necessarily have a strong impact on the ML results.

Mondrian is the only anonymisation method for which we can easily compare the results with existing literature. Compared to \textcite{Last2014}, similar classification performance can be observed for the \textsc{cmc} and \textsc{mgm} datasets. For the \textsc{adult} dataset, the differences are much larger, as Mondrian performs significantly worse in their study. This might be due to differences in the data preparation processes (see \autoref{sub:datasetprep}).  

We have shown that there exists a strong dependency between generalisation and the achievable classification performance especially when the generalised features are strongly correlated with the target variable to be predicted during classification. For suppression, the interpretation is more difficult. We could not find a clear dependency of allowed suppression levels and classification performance. Allowing a certain degree of suppression seems to be advisable, as it allows for more flexibility during the anonymisation step and may reduce fluctuations introduced by generalisation.

Overall, we observe that even for very large $k$ of up to 100 (which is far higher than the values used in practice nowadays), the performance losses remain within acceptable limits. This is of course dataset-dependent, \eg\ the \textsc{adult} dataset exhibits almost no decrease (when using RF and Mondrian) and for \textsc{cmc} with SVM, the loss is only around 7\%. For the datasets \textsc{cahousing} and \textsc{mgm} the loss is somewhat larger (approximately 13\%).

Our investigation represents a first starting point for further analysis of the effects of anonymisation strategies on downstream classification (and potentially other ML) tasks. To foster further research in this direction (\eg\ using larger and more diverse datasets, more classifiers and additional anonymisation techniques), we make all of our our code and resources publicly available, including datasets, method implementations, evaluation code and metadata (such as the employed QIDs and VGHs). More concrete open topics and future research directions are detailed in the following to stimulate further research in this direction.

\paragraph{Future Work}

There are several possibilities for expanding our work. First, the utilised datasets (as listed in \autoref{sub:datasets}) could be extended by additional large-scale real-world datasets. As only relatively few larger datasets (suitable for applying $k$-anonymity algorithms) are publicly available, most research in this area is based on the same datasets; including additional datasets with other characteristics and distributions seems highly desirable. While we already applied our algorithms and analyses on four distinct datasets, acquiring additional datasets (potentially including synthetic datasets) to gain further insight regarding the observations described in \autoref{sec:results} would greatly enhance our work and help identify data-independent behaviour.

Another possibility for improvement is the inclusion of more $k$-anonymity algorithms. In order to cover more distinct approaches to achieve $k$-anonymity, we would in the future like to include algorithms utilising subtree generalisation, such as Top-Down Specialization~\parencite{Fung2005,Fung2007} or $k$-Optimize~\parencite{BayardoAgrawal2005}. Furthermore, a systematic comparison of the effects of different anonymisation strat\-e\-gies (including microaggregation and bucketisation in addition to generalisation and suppression) would be extremely desirable. In particular, as indicated by \textcite{rodriguez2019assessing}, the MDAV algorithm~\parencite{hundepool2003argus} (based on multivariate fixed-size microaggregation~\parencite{domingo2002practical}) might have less impact on classification results. Although we could not confirm this advantage of microaggregation in our preliminary experiments using the $\mu$-ANT library, which implements a variant of MDAV, future experiments should consider other variants of the MDAV algorithm and investigate this question in further detail.

During our research we found that, unfortunately, both the implemented algorithms and the results (and sometimes even the underlying data) are in many cases either not accessible at all or at least not easily accessible. We hope that our findings as well as providing our source code and data might lay the groundwork for further projects and help promote open science approaches, in general. Our source code and the utilised datasets are publicly available at \href{https://github.com/fhstp/k-AnonML}{github.com/fhstp/k-AnonML}.

\section*{Acknowledgements}
This research was funded by the Austrian Research Promotion Agency (FFG) through COIN project 866880 ``Big Data Analytics''. The financial support by the Austrian Research Promotion Agency and the Federal Ministry for Digital and Economic Affairs is gratefully acknowledged.

\printbibliography

\newpage
\appendix
\section{Supplementary Figures}
\setcounter{figure}{0}    

\begin{figure*}[h!]
    \centering
    \includegraphics[width=\linewidth]{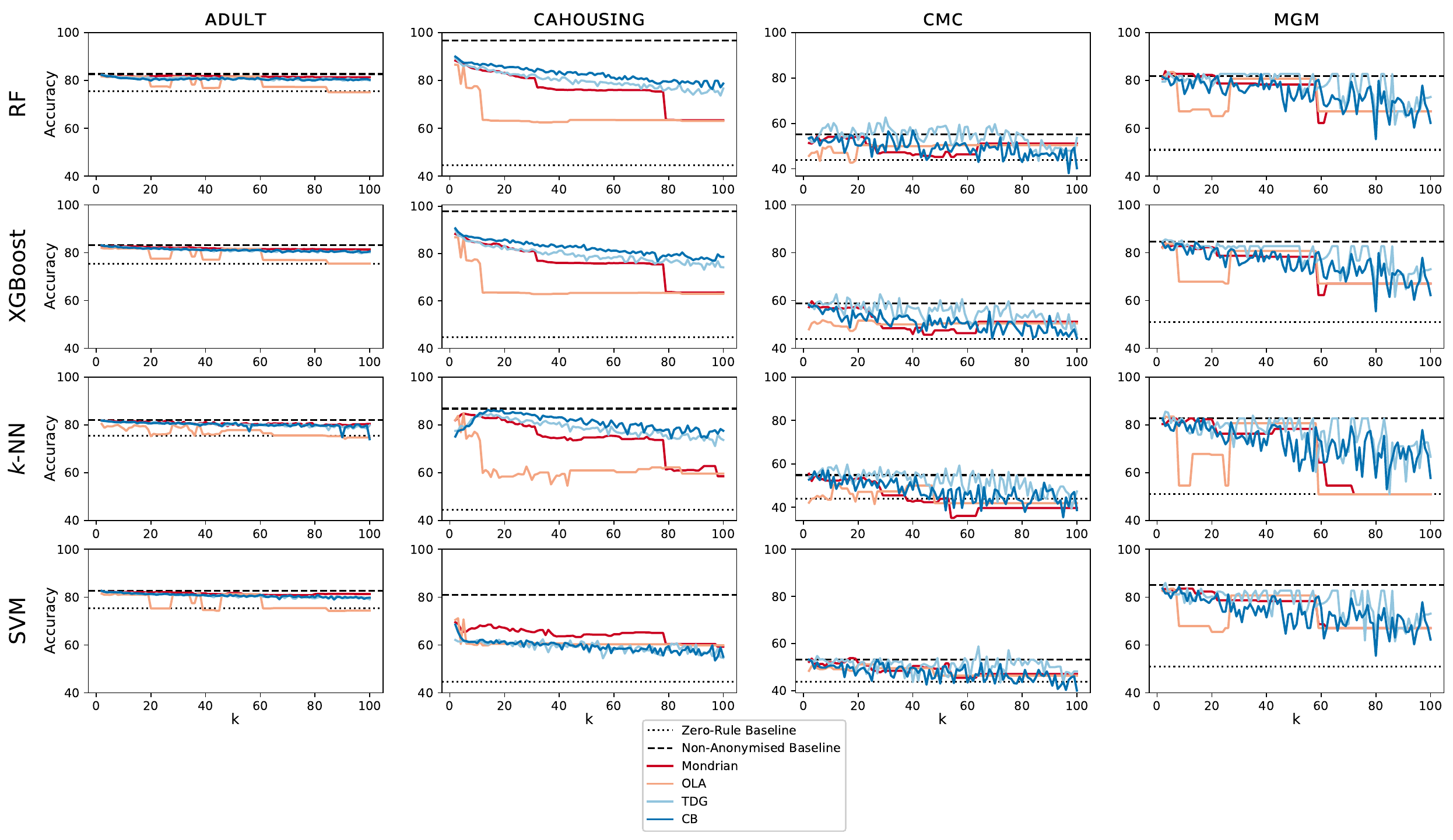}
    \caption{Overview of the performance in terms of classification accuracy for all four datasets, all four classifiers and all anonymisation methods for $k$ ranging from 2 to 100.}
    \label{fig:overviewAccuracy}
\end{figure*}

\begin{figure*}[h!]
    \centering
    \includegraphics[width=\linewidth]{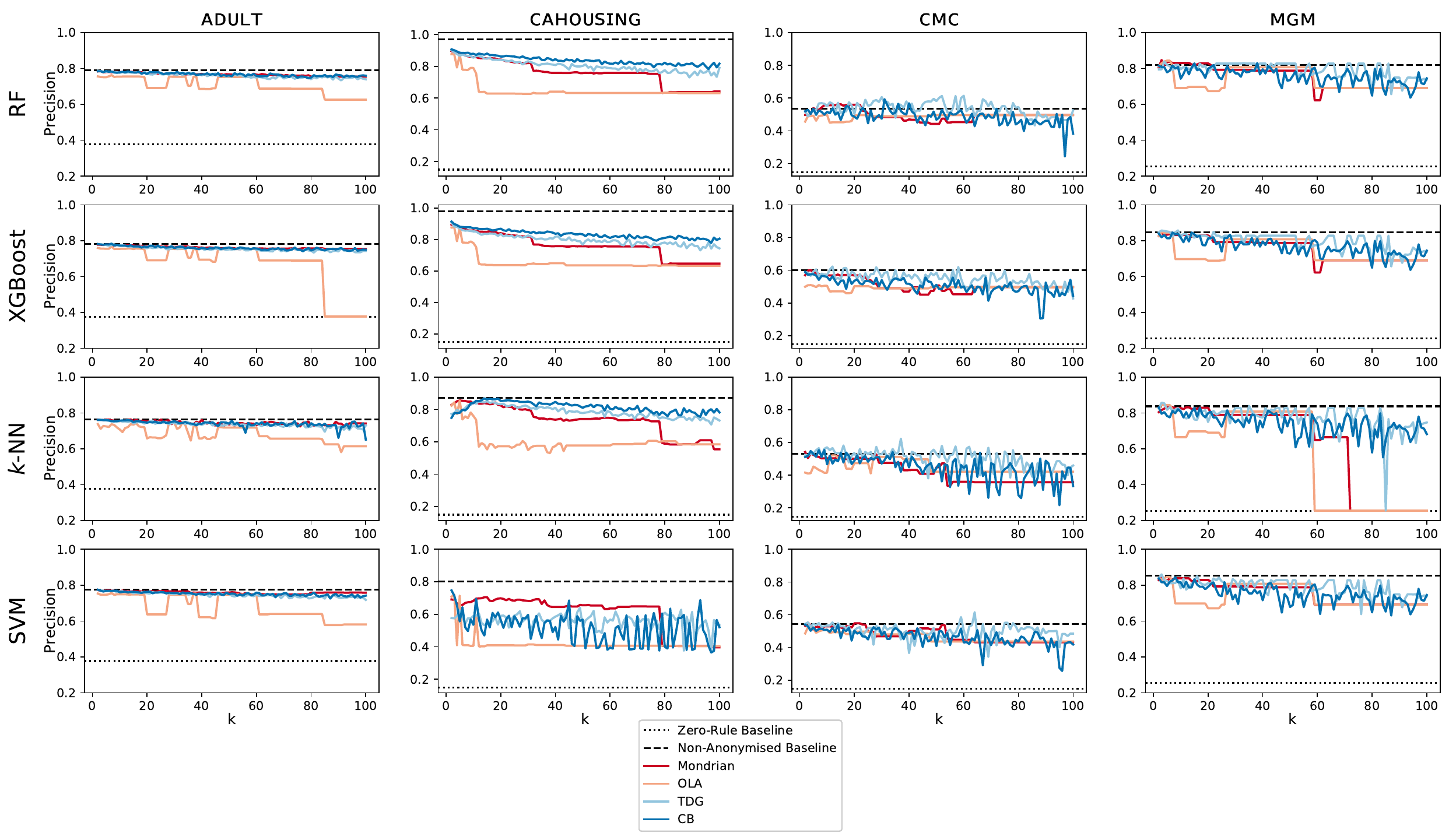}
    \caption{Overview of the performance in terms of precision for all four datasets, all four classifiers and all anonymisation methods for $k$ ranging from 2 to 100.}
    \label{fig:overviewPrecision}
\end{figure*}

\begin{figure*}[h!]
    \centering
    \includegraphics[width=\linewidth]{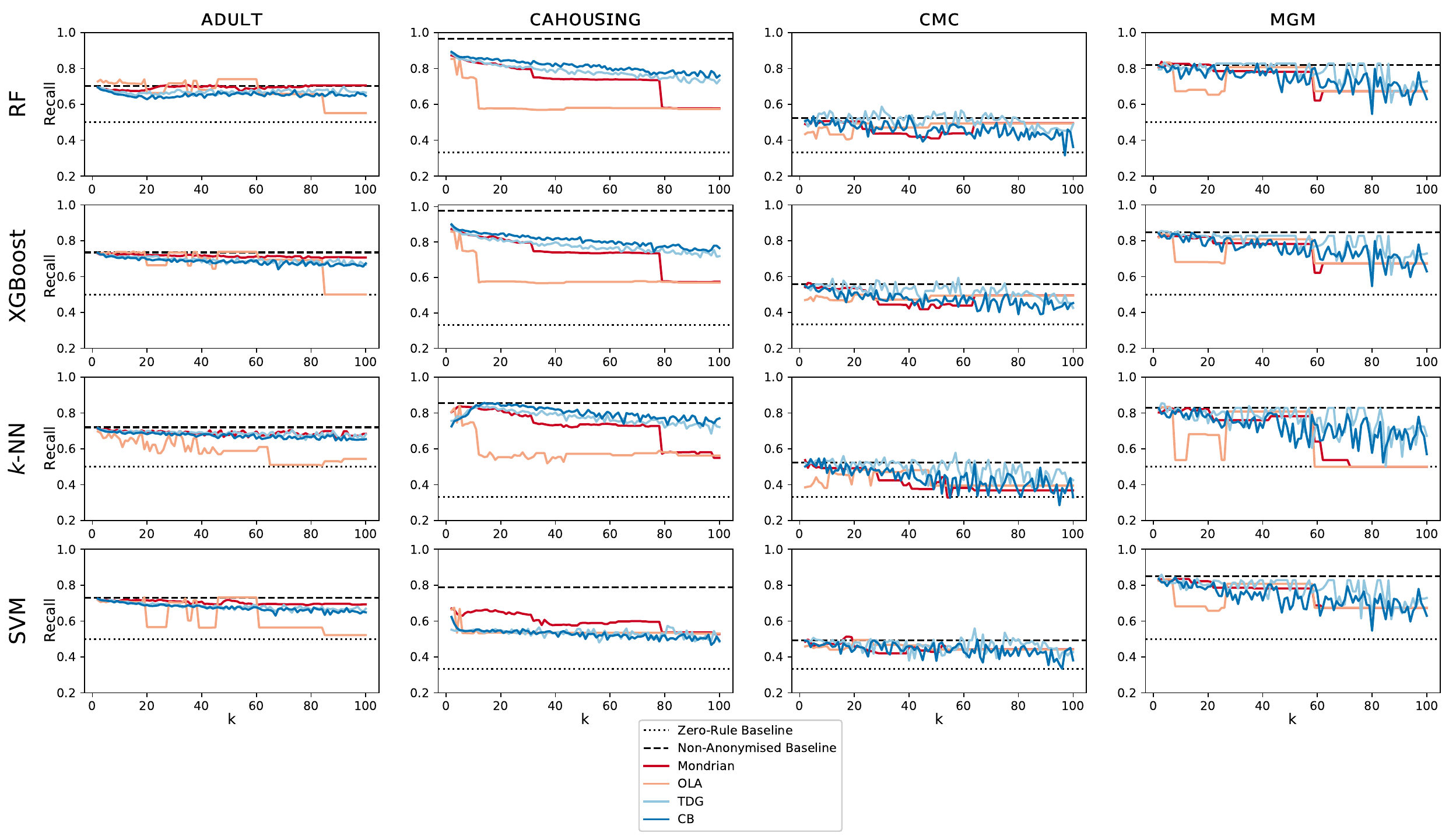}
    \caption{Overview of the performance in terms of recall for all four datasets, all four classifiers and all anonymisation methods for $k$ ranging from 2 to 100.}
    \label{fig:overviewRecall}
\end{figure*}

\begin{figure*}[h!]
    \centering
    \includegraphics[width=\linewidth]{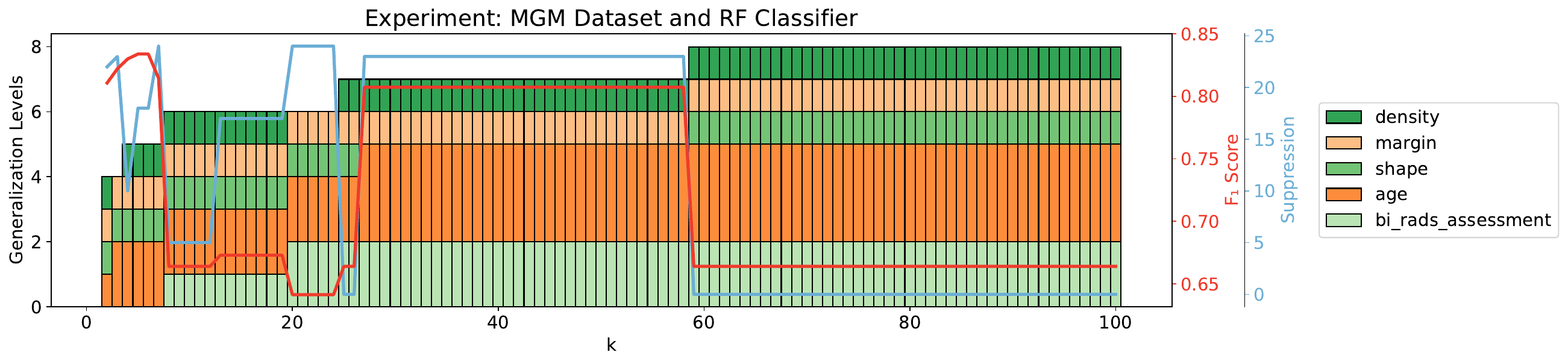}
    \caption{Analysis of the combined effects of generalisation and suppression for OLA on the \textsc{mgm} dataset. Each coloured bar corresponds to a feature (\eg\ \texttt{density}) and its height to the generalisation level (in the generalisation hierarchy). The green curve represents the number of suppressed records and the blue curve the F\textsubscript{1}~score obtained for various $k$ values.}
    \label{fig:gen_sup_mgm}
\end{figure*}

\begin{figure*}[!ht]
    \centering
    \includegraphics[width=\linewidth]{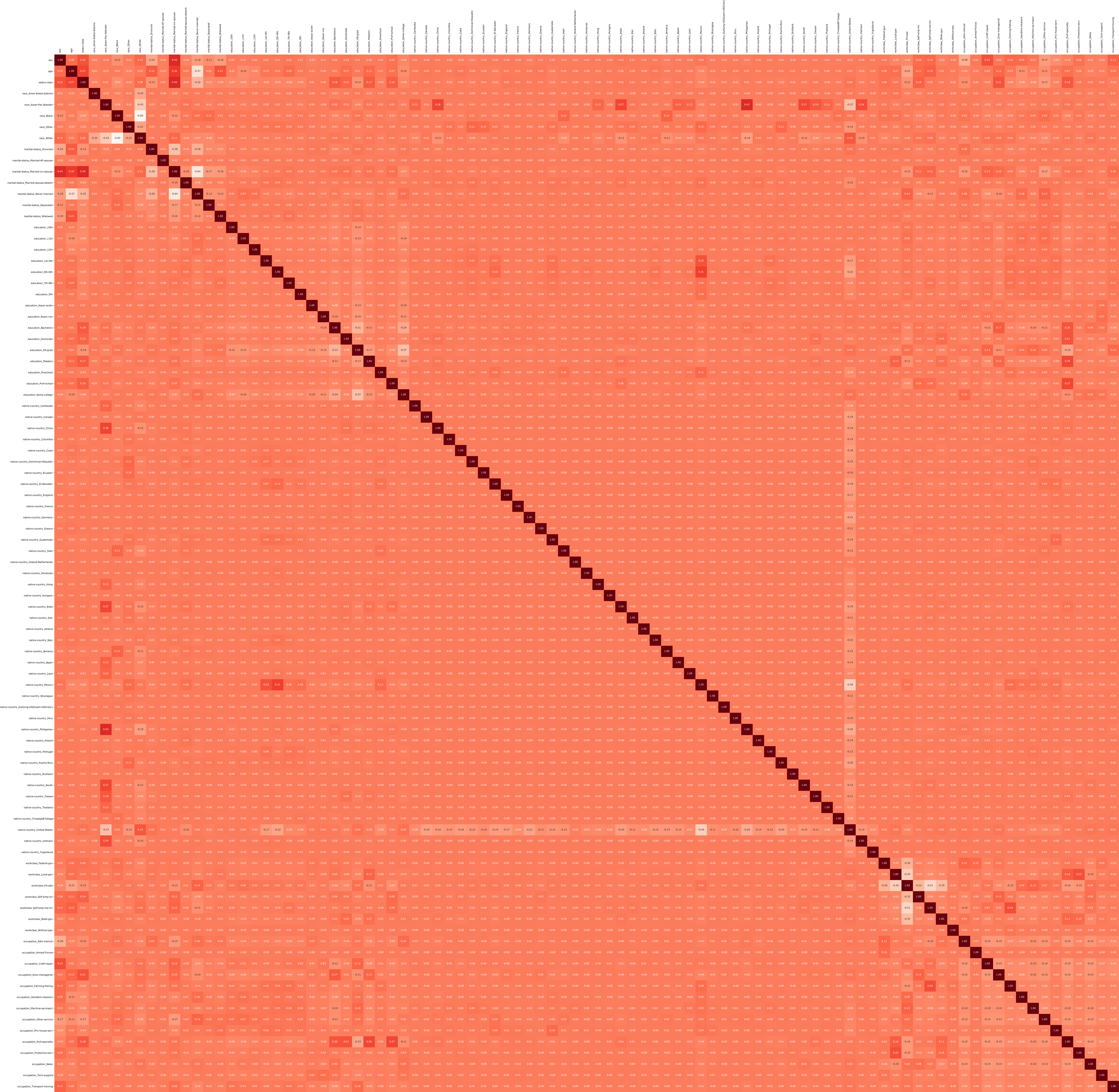}
    \caption{Correlation matrix for the \textsc{adult} dataset, showing Spearman's rank correlation coefficients between the target variable and the features. (Categorical features were one-hot encoded.)}
    \label{fig:corr_matrix_adult}
\end{figure*}

\begin{figure*}[!ht]
    \centering
    \includegraphics[]{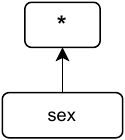}
    \caption{Generalisation hierarchy for the \texttt{sex} attribute in the \textsc{adult} dataset.}
    \label{fig:adult_sex}
\end{figure*}

\begin{figure*}[!ht]
    \centering
    \includegraphics[width=\linewidth]{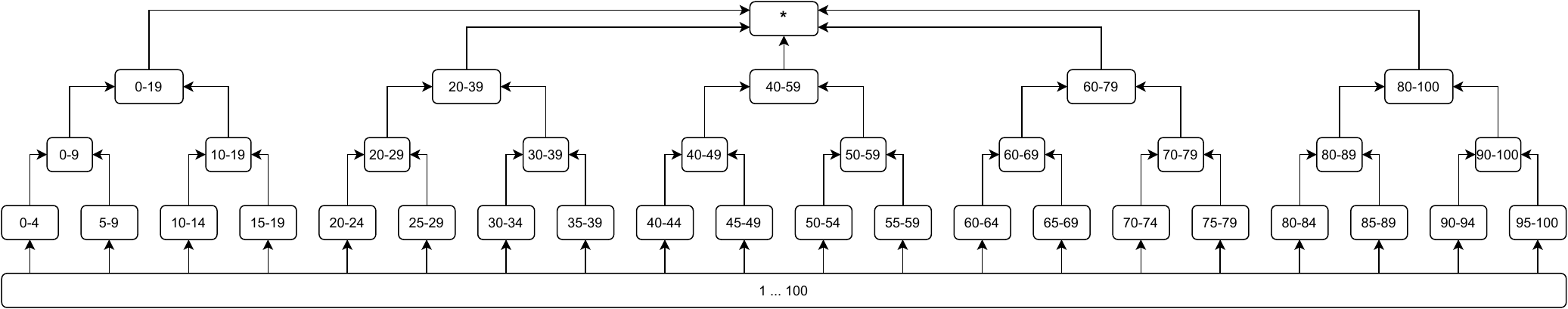}
    \caption{Generalisation hierarchy for the \texttt{age} attribute in the \textsc{adult} dataset.}
    \label{fig:adult_age}
\end{figure*}

\begin{figure*}[!ht]
    \centering
    \includegraphics[]{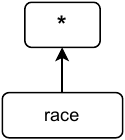}
    \caption{Generalisation hierarchy for the \texttt{race} attribute in the \textsc{adult} dataset.}
    \label{fig:adult_race}
\end{figure*}

\begin{figure*}[!ht]
    \centering
    \includegraphics[]{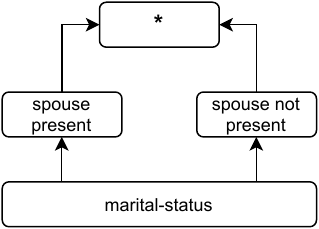}
    \caption{Generalisation hierarchy for the \texttt{marital-status} attribute in the \textsc{adult} dataset.}
    \label{fig:adult_marital-status}
\end{figure*}

\begin{figure*}[!ht]
    \centering
    \includegraphics[]{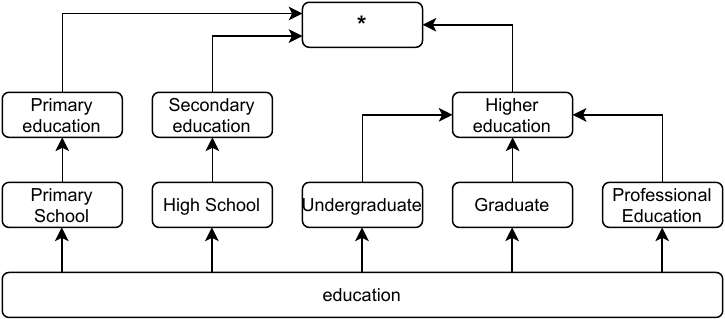}
    \caption{Generalisation hierarchy for the \texttt{education} attribute in the \textsc{adult} dataset.}
    \label{fig:adult_education}
\end{figure*}

\begin{figure*}[!ht]
    \centering
    \includegraphics[]{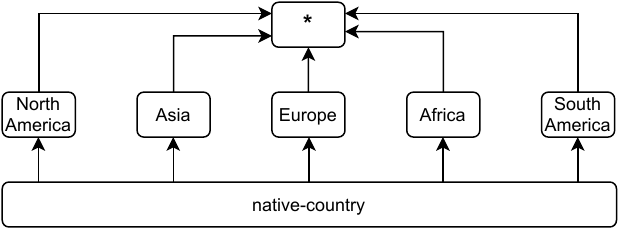}
    \caption{Generalisation hierarchy for the \texttt{native-country} attribute in the \textsc{adult} dataset.}
    \label{fig:adult_country}
\end{figure*}

\begin{figure*}[!ht]
    \centering
    \includegraphics[]{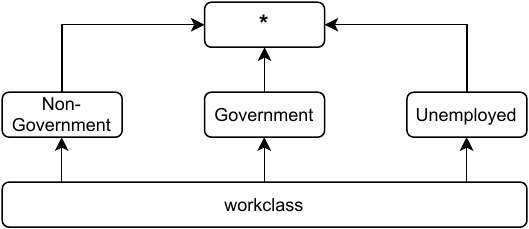}
    \caption{Generalisation hierarchy for the \texttt{workclass} attribute in the \textsc{adult} dataset.}
    \label{fig:adult_workclass}
\end{figure*}

\begin{figure*}[!ht]
    \centering
    \includegraphics[]{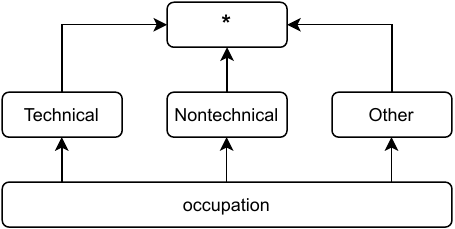}
    \caption{Generalisation hierarchy for the \texttt{occupation} attribute in the \textsc{adult} dataset.}
    \label{fig:adult_occupation}
\end{figure*}

\begin{figure*}[!ht]
    \centering
    \includegraphics[]{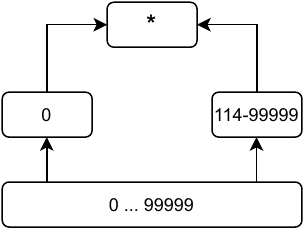}
    \caption{Generalisation hierarchy for the \texttt{capital-gain} attribute in the \textsc{adult} dataset.}
    \label{fig:adult_capital}
\end{figure*}

\begin{figure*}[!ht]
    \centering
    \includegraphics[]{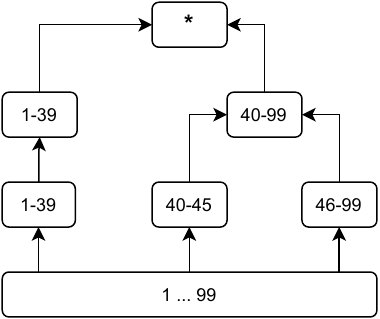}
    \caption{Generalisation hierarchy for the \texttt{hours-per-week} attribute in the \textsc{adult} dataset.}
    \label{fig:adult_hours}
\end{figure*}

\begin{figure*}[!ht]
    \centering
    \includegraphics[width=\linewidth]{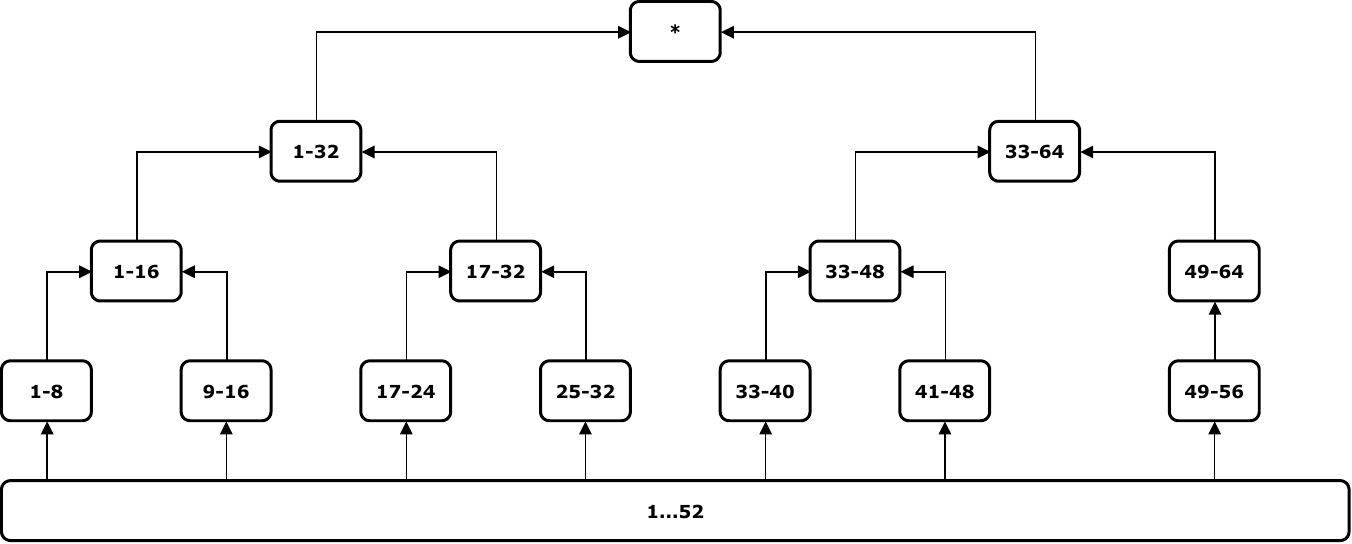}
    \caption{Generalisation hierarchy for the \texttt{housing\_median\_age} attribute in the \textsc{cahousing} dataset.}
    \label{fig:cah_age}
\end{figure*}

\begin{figure*}[!ht]
    \centering
    \includegraphics[width=\linewidth]{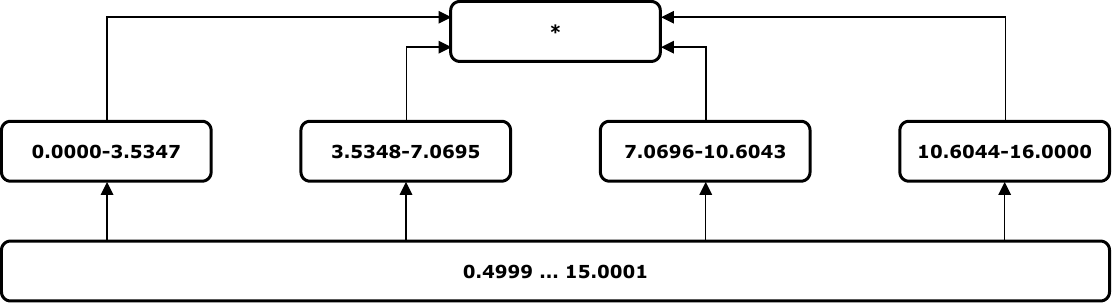}
    \caption{Generalisation hierarchy for the \texttt{median\_income} attribute in the \textsc{cahousing} dataset.}
    \label{fig:cah_income}
\end{figure*}

\begin{figure*}[!ht]
    \centering
    \includegraphics[]{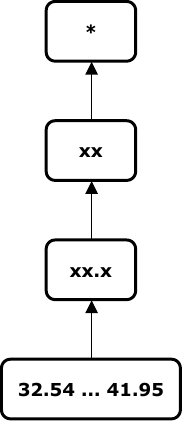}
    \caption{Generalisation hierarchy for the \texttt{latitude} attribute in the \textsc{cahousing} dataset.}
    \label{fig:cah_lat}
\end{figure*}

\begin{figure*}[!ht]
    \centering
    \includegraphics[]{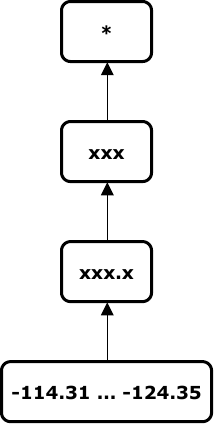}
    \caption{Generalisation hierarchy for the \texttt{longitude} attribute in the \textsc{cahousing} dataset.}
    \label{fig:cah_lon}
\end{figure*}

\begin{figure*}[!ht]
    \centering
    \includegraphics[width=\linewidth]{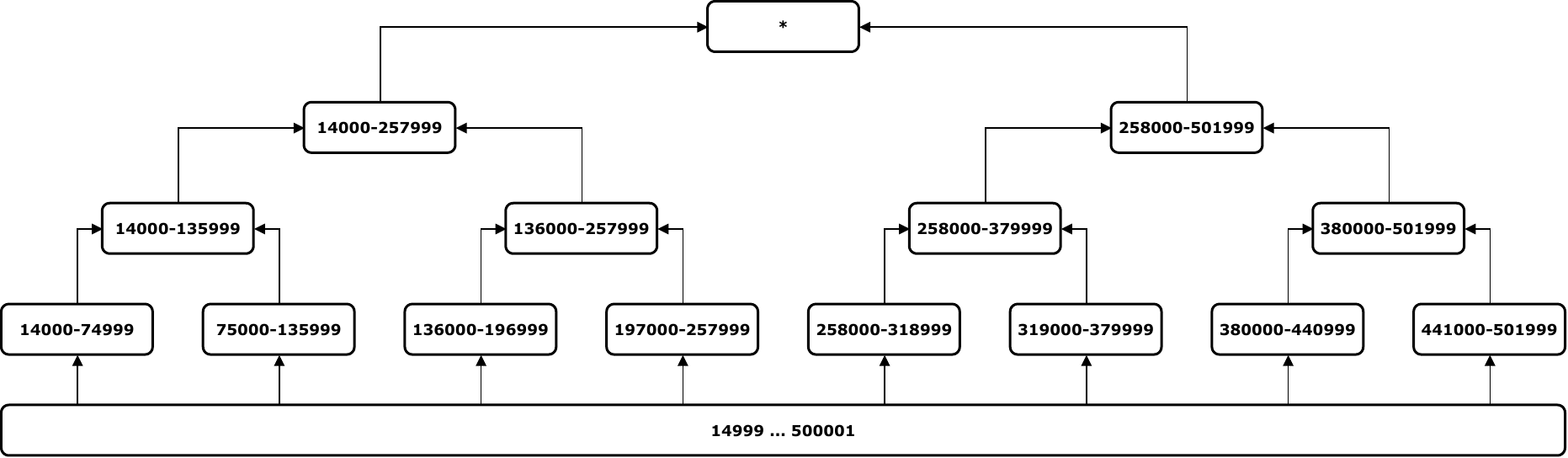}
    \caption{Generalisation hierarchy for the \texttt{median\_house\_value} attribute in the \textsc{cahousing} dataset.}
    \label{fig:cah_value}
\end{figure*}

\begin{figure*}[!ht]
    \centering
    \includegraphics[width=\linewidth]{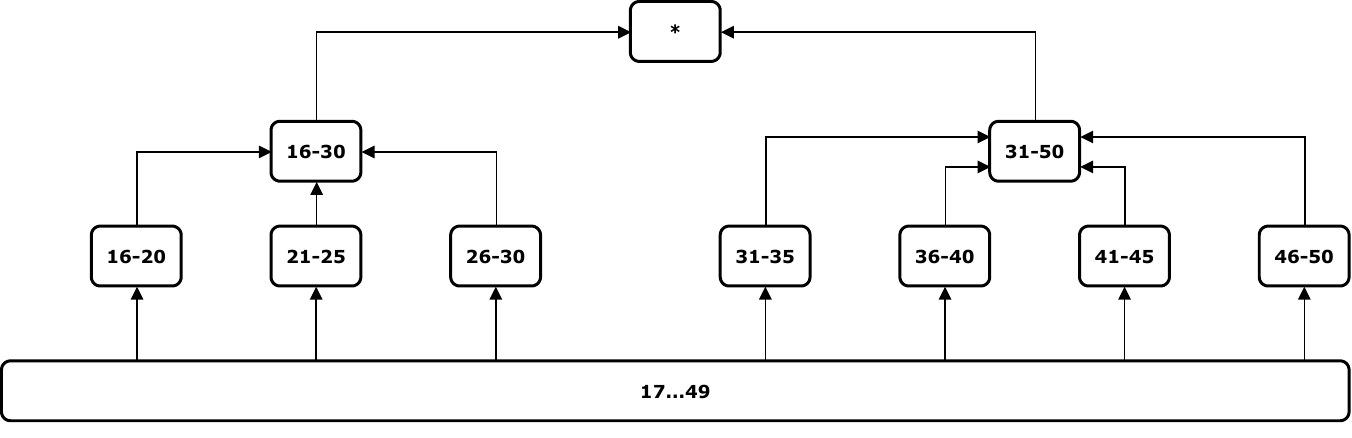}
    \caption{Generalisation hierarchy for the \texttt{wife\_age} attribute in the \textsc{cmc} dataset.}
    \label{fig:cmc_age}
\end{figure*}

\begin{figure*}[!ht]
    \centering
    \includegraphics[width=\linewidth]{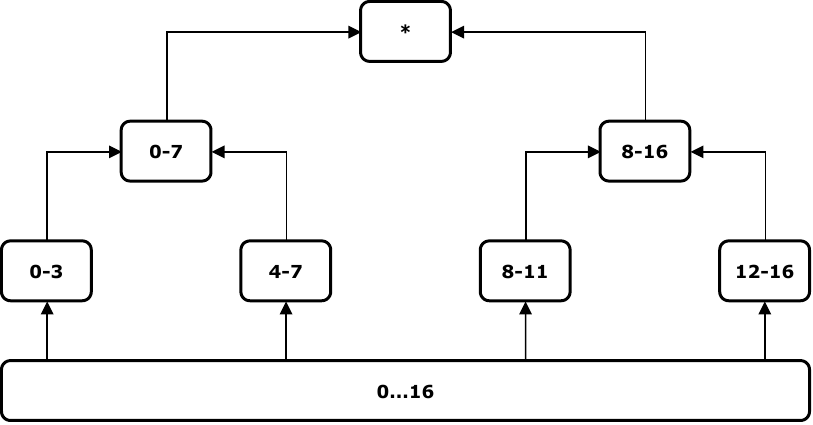}
    \caption{Generalisation hierarchy for the \texttt{num\_children} attribute in the \textsc{cmc} dataset.}
    \label{fig:cmc_child}
\end{figure*}

\begin{figure*}[!ht]
    \centering
    \includegraphics[]{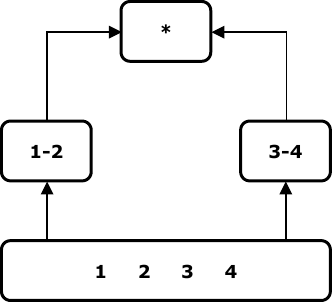}
    \caption{Generalisation hierarchy for the \texttt{wife\_edu} attribute in the \textsc{cmc} dataset.}
    \label{fig:cmc_edu}
\end{figure*}

\begin{figure*}[!ht]
    \centering
    \includegraphics[width=\linewidth]{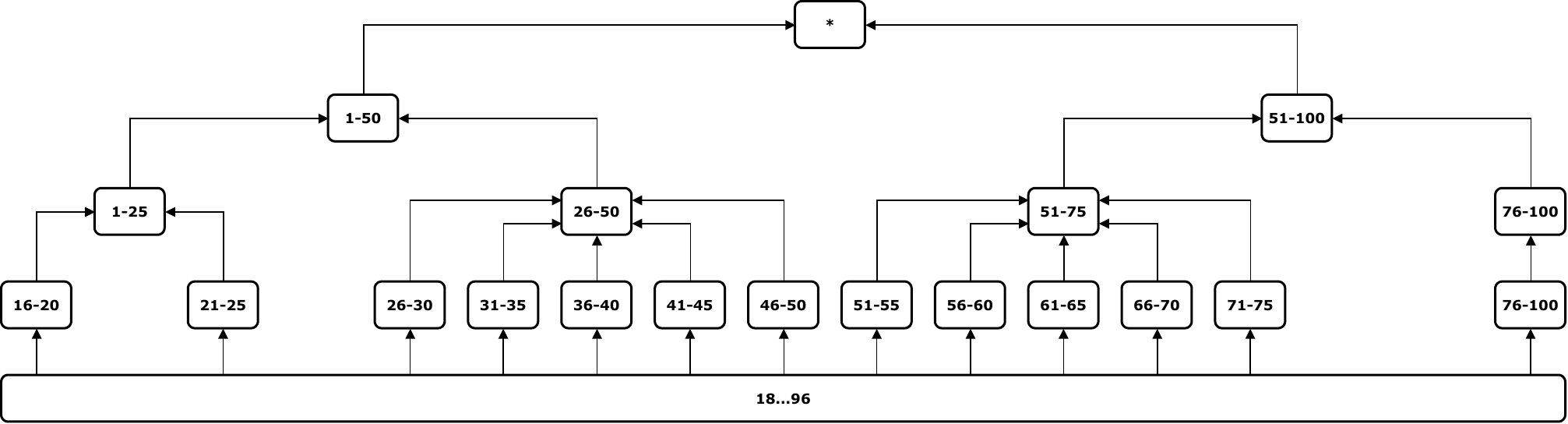}
    \caption{Generalisation hierarchy for the \texttt{age} attribute in the \textsc{mgm} dataset.}
    \label{fig:mgm_age}
\end{figure*}

\begin{figure*}[!ht]
    \centering
    \includegraphics[]{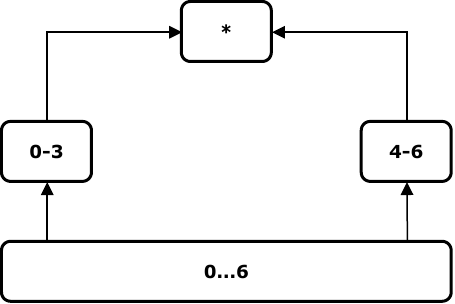}
    \caption{Generalisation hierarchy for the \texttt{bi\_rads\_assessment} attribute in the \textsc{mgm} dataset.}
    \label{fig:mgm_birads}
\end{figure*}

\begin{figure*}[!ht]
    \centering
    \includegraphics[]{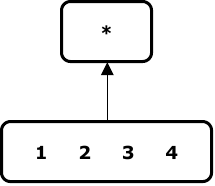}
    \caption{Generalisation hierarchy for the \texttt{density} attribute in the \textsc{mgm} dataset.}
    \label{fig:mgm_dens}
\end{figure*}

\begin{figure*}[!ht]
    \centering
    \includegraphics[]{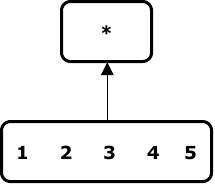}
    \caption{Generalisation hierarchy for the \texttt{margin} attribute in the \textsc{mgm} dataset.}
    \label{fig:mgm_margin}
\end{figure*}

\begin{figure*}[!ht]
    \centering
    \includegraphics[]{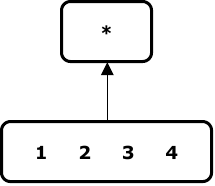}
    \caption{Generalisation hierarchy for the \texttt{shape} attribute in the \textsc{mgm} dataset.}
    \label{fig:mgm_shape}
\end{figure*}
\end{document}